\documentclass{article}

\usepackage[utf8]{inputenc} %
\usepackage[T1]{fontenc}    %
\usepackage{microtype}
\usepackage{graphicx}
\usepackage{booktabs} %
\usepackage{hyperref}
\usepackage{setspace}
\usepackage{url}            %
\usepackage{booktabs}       %
\usepackage{amsfonts}       %
\usepackage{nicefrac}       %
\usepackage[usenames,dvipsnames]{xcolor}
\usepackage{paralist}
\usepackage{placeins}
\usepackage{float}
\usepackage{wrapfig}
\usepackage{siunitx}
\usepackage{soul}
\usepackage{realboxes}

\usepackage{hyperref}

\newcommand{\changed}[1]{#1}

\usepackage[accepted]{icml2021}

\icmltitlerunning{Scalable Marginal Likelihood Estimation for Model Selection in Deep Learning}

\usepackage{microtype}
\usepackage{graphicx}
\usepackage{float}
\usepackage{subcaption}
\usepackage{color}
\usepackage{url}
\usepackage{rotating}
\usepackage{adjustbox}
\usepackage{siunitx}
\usepackage{etoolbox}
\usepackage{booktabs}
\usepackage{multirow}
\usepackage{multicol}
\usepackage{array}
\usepackage{enumitem}
\newcolumntype{C}{>{$}c<{$}}
\newcolumntype{L}{>{$}l<{$}}
\newcolumntype{R}{>{$}r<{$}}

\usepackage{amsmath}
\usepackage{amsfonts}
\usepackage{bbm}
\usepackage{mathtools}
\usepackage{xspace}
\usepackage{nicefrac}

\usepackage[capitalise,nameinlink]{cleveref}
\crefname{section}{Sec.}{Secs.}
\crefname{appendix}{App.}{Apps.}
\crefname{algorithm}{Alg.}{Algs.}
\creflabelformat{equation}{#2\textup{#1}#3}  %

\usepackage{tikz}
\usetikzlibrary{positioning}
\usetikzlibrary{calc}
\usetikzlibrary{arrows.meta}
\usetikzlibrary{backgrounds}
\usetikzlibrary{decorations.pathmorphing, patterns}
\usetikzlibrary{decorations.pathreplacing}
\usetikzlibrary{plotmarks}

\usepackage{contour}

\definecolor{lightgray}{gray}{0.85}
\definecolor{lightlightgray}{gray}{0.9}
\definecolor{C1}{HTML}{1F77B4}
\definecolor{C2}{HTML}{FF7F0E}
\definecolor{C3}{HTML}{2CA02C}
\definecolor{C4}{HTML}{D62728}
\definecolor{C5}{HTML}{9467BD}
\colorlet{C1light}{C1!70!white}
\colorlet{C2light}{C2!70!white}
\colorlet{C3light}{C3!70!white}
\colorlet{C4light}{C4!70!white}
\colorlet{C5light}{C5!70!white}
\colorlet{C1lighter}{C1!40!white}
\colorlet{C2lighter}{C2!40!white}
\colorlet{C3lighter}{C3!40!white}
\colorlet{C4lighter}{C4!40!white}
\colorlet{C5lighter}{C5!40!white}
\colorlet{C1vlight}{C1!20!white}
\colorlet{C2vlight}{C2!20!white}
\colorlet{C3vlight}{C3!20!white}
\colorlet{C4vlight}{C4!20!white}
\colorlet{C5vlight}{C5!20!white}

\usepackage[normalem]{ulem}

\newcommand{\param}{\ensuremath{\vtheta}\xspace}

\newcommand{\paramstar}{{\vtheta_\ast}}
\newcommand{\model}{\ensuremath{\sM}\xspace}
\newcommand{\modelmap}{\ensuremath{\sM_\ast}\xspace}
\newcommand{\hypothesis}{\sM}

\newcommand{\prior}{\ensuremath{p(\vtheta)}\xspace}
\newcommand{\priorhyp}{\ensuremath{p(\vtheta \given \hypothesis)}\xspace}

\newcommand{\likelihoodhyp}{\ensuremath{p(\data \given \param, \hypothesis)}\xspace}
\newcommand{\likelihoodn}{\ensuremath{p(\vy_n \given \vf(\vx_n, \param))}\xspace}

\newcommand{\onelikelihoodf}{\ensuremath{p(\vy\given \vf)}\xspace}

\newcommand{\ggn}{\textsc{ggn}\xspace}

\newcommand{\adam}{\textsc{adam}\xspace}
\newcommand{\kfac}{\textsc{kfac}\xspace}
\newcommand{\ef}{\textsc{ef}\xspace}

\newcommand{\ReLU}{\texttt{ReLU}\xspace}

\newcommand{\jac}{\ensuremath{\mJ_{\vtheta}(\vx)}\xspace}
\newcommand{\jacn}{\ensuremath{\mJ_{\vtheta}(\vx_n)}\xspace}

\newcommand{\iid}{\emph{i.i.d.}}
\newcommand{\pms}[1]{\ensuremath{{\scriptstyle\pm #1}}}

\newcommand{\transpose}{^\mathrm{\textsf{\tiny T}}}
\newcommand{\diag}{\operatorname{diag}}
\newcommand{\kron}{\otimes}

\newcommand{\inv}{^{-1}}

\DeclareMathOperator*{\argmax}{arg\,max}

\newcommand{\Var}{\mathrm{Var}}

\newcommand{\R}{\mathbb{R}}

\def\sD{{\mathcal{D}}}

\def\sM{{\mathcal{M}}}

\def\D{{\sD}}
\def\data{{\sD}}

\def\vzero{{\mathbf{0}}}

\def\vf{{\mathbf{f}}}
\def\vg{{\mathbf{g}}}

\def\vq{{\mathbf{q}}}

\def\vw{{\mathbf{w}}}
\def\vx{{\mathbf{x}}}
\def\vy{{\mathbf{y}}}

\def\vtheta{{\boldsymbol{\theta}\xspace}}
\def\vmu{{\boldsymbol{\mu}\xspace}}

\def\mG{{\mathbf{G}}}
\def\mH{{\mathbf{H}}}
\def\mI{{\mathbf{I}}}
\def\mJ{{\mathbf{J}}}

\def\mL{{\mathbf{L}}}
\def\mM{{\mathbf{M}}}

\def\mP{{\mathbf{P}}}
\def\mQ{{\mathbf{Q}}}

\def\mW{{\mathbf{W}}}

\def\mLambda{{\boldsymbol{\Lambda}\xspace}}

\newcommand{\gauss}{\mathcal{N}}

\newcommand{\softmax}{\mathrm{softmax}}

\newcommand{\half}{\mbox{$\frac{1}{2}$}}

\newcommand{\given}{\mbox{$|$}}

\newcommand{\la}{\mbox{$\leftarrow$}}

\newcommand{\order}[1]{\ensuremath{\mathcal{O}(#1)}}

\newcommand{\sqr}[1]{\left[#1\right]}

\newcommand{\tikzunderlinewithtext}[3]{%
    \contourlength{0.08em}
    \tikz[baseline=(tounderline.base)]{
        \node[inner sep=0pt,outer sep=0pt] (tounderline) {\contour{white}{#2}};
        \useasboundingbox ($(tounderline.base west)+(0, 0)$) rectangle (tounderline.base east);
        \begin{scope}[on background layer]
            \draw[#1] ($(tounderline.base west)-(0,0.1)$) -- ($(tounderline.base east)-(0,0.1)$) node[pos=0.5, below] {\textcolor{black}{#3}};
        \end{scope}
    }%
}%
\newcommand{\mathunderlinewithtext}[2]{
    \tikzunderlinewithtext{decorate,decoration={brace,mirror}, line width=0.5pt, black}{$#1$}{{\scriptsize $#2$}}}

\usepackage[disable]{todonotes}
\makeatletter
\newcommand*\iftodonotes{\if@todonotes@disabled\expandafter\@secondoftwo\else\expandafter\@firstoftwo\fi}  %
\makeatother

\definecolor{dandelion}{HTML}{FFD464}

\makeatletter
\renewcommand{\paragraph}{%
  \@startsection{paragraph}{4}%
  {\z@}{0.25ex \@plus 0.5ex \@minus .2ex}{-1em}%
  {\normalfont\normalsize\bfseries}%
}
\makeatother

\begin{document}

\twocolumn[
\icmltitle{Scalable Marginal Likelihood Estimation for Model Selection in Deep Learning}

\icmlsetsymbol{equal}{*}
\icmlsetsymbol{dmnote}{$\dagger$}

\begin{icmlauthorlist}
\icmlauthor{Alexander Immer}{eth,cls}
\icmlauthor{Matthias Bauer}{dmnote,mpi,cam}
\icmlauthor{Vincent Fortuin}{eth}
\icmlauthor{Gunnar R\"atsch}{eth,cls}
\icmlauthor{Mohammad Emtiyaz Khan}{rik}
\end{icmlauthorlist}

\icmlaffiliation{eth}{Department of Computer Science, ETH Zurich, Switzerland}
\icmlaffiliation{cls}{Max Planck ETH Center for Learning Systems (CLS)}
\icmlaffiliation{mpi}{Max Planck Institute for Intelligent Systems, Germany}
\icmlaffiliation{cam}{University of Cambridge, UK}
\icmlaffiliation{rik}{RIKEN Center for Advanced Intelligence Project, Japan}

\icmlcorrespondingauthor{Alexander Immer}{alexander.immer@inf.ethz.ch}

\icmlkeywords{Machine Learning, ICML, Bayesian Deep Learning, Marginal Likelihood, Empirical Bayes, Gauss-Newton, Fisher Information}

\vskip 0.3in
]

\printAffiliationsAndNotice{}{\mbDMnote}

\begin{abstract}
Marginal-likelihood based model-selection, even though promising, is rarely used in deep learning due to estimation difficulties.
Instead, most approaches rely on validation data, which may not be readily available.~In this work, we present a scalable marginal-likelihood estimation method to select \changed{both hyperparameters and network architectures,} based on the training data alone.~Some hyperparameters can be estimated online during training, simplifying the procedure.~Our marginal-likelihood estimate is based on Laplace's method and Gauss-Newton approximations to the Hessian, and it outperforms cross-validation and manual-tuning on standard regression and image classification datasets, especially in terms of calibration and out-of-distribution detection.
Our work shows that marginal likelihoods can improve generalization and be useful when validation data is unavailable (e.g., in nonstationary settings).

\end{abstract}

\section{Introduction}
\label{sec:intro}
Bayesian deep learning has made great strides on approximate inference but little has been done regarding model selection.
Bayesian predictive distributions have been used to improve calibration, detect out-of-distribution data, and sometimes even improve the accuracy~\citep{osawa2019practical, maddox2019simple}.
To that end, a wide variety of scalable approximate inference methods have been explored, including Laplace's method~\citep{ritter2018scalable}, variational approximations~\citep{blundell2015weight,khan2018fast}, sampling methods~\citep{wenzel2020good}, and ensembles~\citep{lakshminarayanan2017simple}.
Still, there is not much work on model-selection, even though it was one of the original motivations for Bayesian neural networks~\citep{buntine1991bayesian, mackay1995probable, neal1995bayesian}.

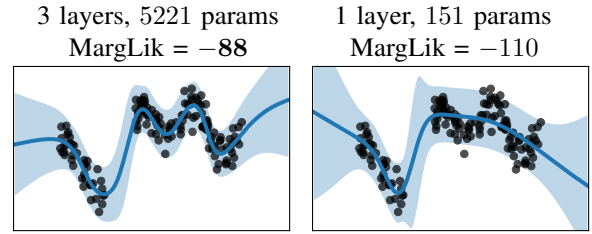
\begin{figure}[t]
  \centering
  \begin{tikzpicture}
    \node (plot){\includegraphics[width=\columnwidth]{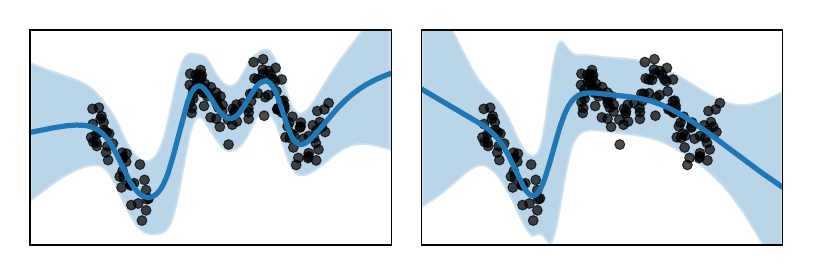}};
  \node at ($(plot.north)+(-1.9,0.0)$) {
      \begin{tikzpicture}
        \node [text width=3.5cm,align=center] {3 layers, $5221$ params MargLik = $\mathbf{-88}$};
      \end{tikzpicture}
  };
  \node at ($(plot.north)+(1.9,0.0)$) {
      \begin{tikzpicture}
        \node [text width=3cm,align=center] {1 layer, $151$ params MargLik = $-110$};
      \end{tikzpicture}
  };
  \end{tikzpicture}
  \vskip -1.5em
  \caption{
    A deeper 3-layer network \textit{(left)} has a better marginal likelihood compared to a 1-layer network \textit{(right)}. This agrees with the fit where the deeper network appears to explain the `sinusoidal' trend better. %
    We optimize the prior and noise variance on marginal likelihood estimates \emph{during training} (see \cref{alg:online_marglik} and \cref{sec:toy_examples} for details).
    }
  \vskip -1em
  \label{fig:toy_model_selection}
\end{figure}

\begin{figure*}[ht]
     \vspace{-0.5em}
      \begin{subfigure}{0.33\textwidth}
        \centerline{\includegraphics[scale=0.92]{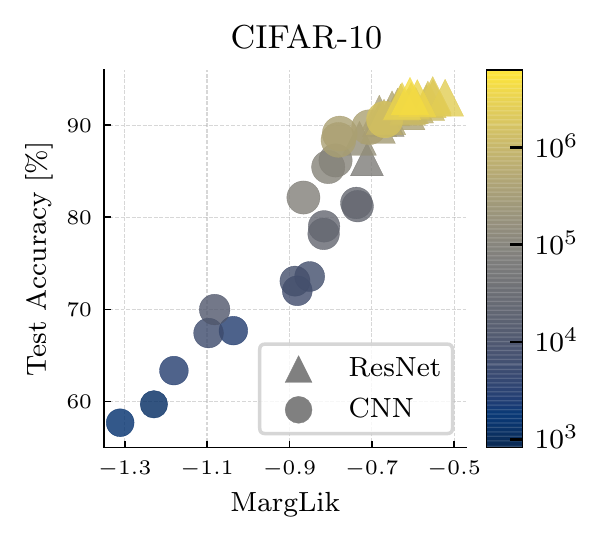}}
                    \vspace{-.5em}
      \label{fig:cifar10_arch_bw}
      \end{subfigure}
      \begin{subfigure}{0.31\textwidth}
        \centerline{\includegraphics[scale=0.92]{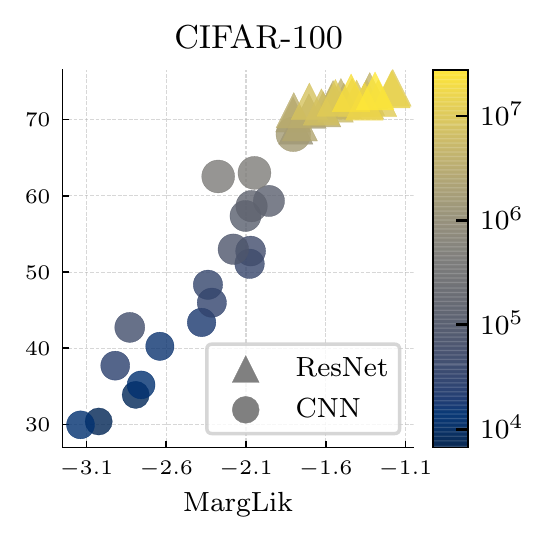}}
                    \vspace{-.5em}
        \label{fig:cifar100_arch_bw}
      \end{subfigure}
      \begin{subfigure}{0.33\textwidth}
        \centerline{\includegraphics[scale=0.92]{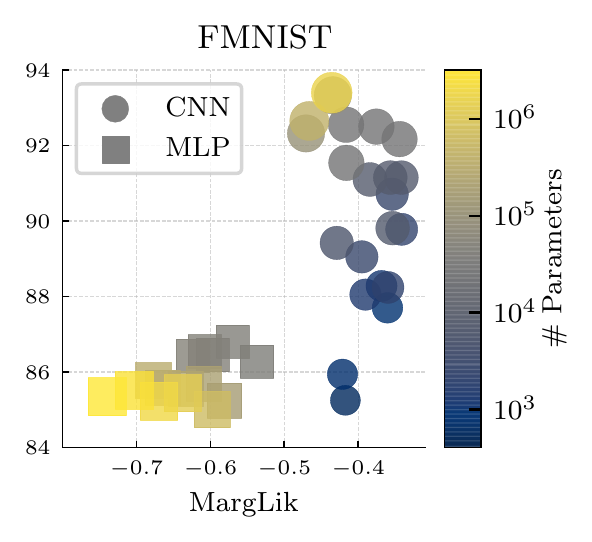}}
                    \vspace{-.5em}
        \label{fig:fmnist_arch_bw}
      \end{subfigure}
      \vspace{-0.5em}
    \caption{Each dot above shows a model of different size and/or architecture (around 40 models per plot of varying widths and depths). Models with higher training marginal-likelihood tend to have higher test accuracy. For similar performance, smaller models tend to have a higher marginal-likelihood as desired. Marker size and color changes with the number of parameters. See \cref{sec:image_classification} for details and \cref{app:exp:model_selection} for a full list of models with accuracy and marginal-likelihood estimates.}
    \label{fig:teaser_marglik_accuracy}
\end{figure*}

Bayesian model-selection uses the \emph{marginal likelihood}---the normalizing constant of the posterior distribution---which can be estimated solely from training data and without any validation data.
Also known as \emph{empirical Bayes} \citep{robbins1955empirical} or \emph{type-II maximum likelihood} learning \citep{rasmussen2006gaussian}, it is closely related to Occam's razor \citep{Jefferys1992_occams_razor, rasmussen2001occam, mackay2003information}.
It can also be seen as a form of cross-validation based on the training data~\citep{fong2020marginal}, and it is commonly used in Gaussian processes (\changed{GPs}) and Deep GPs to select hyperparameters~\citep{rasmussen2006gaussian,deepgp2013} or even learn invariances from data~\citep{van2018learning,deep_gp_image_classification_2020}.
However, it is rarely used in deep learning because it is challenging to estimate \citep{llorente2020marginal}.

\changed{The marginal likelihood is intractable for even medium-sized neural-network models, and most Bayesian deep learning approaches simply use validation data~\citep{khan2018fast, zhang2018noisy, osawa2019practical}.}
Originally, \citet{mackay1995probable} advocated the use of Laplace's method for small-scale neural networks. This requires Hessian computations and does not scale to large problems.
Recently, \citet{khan2019approximate} showed promising results for small-scale regression using a GP viewpoint of deep networks, and \citet{lyle2020bayesian} used infinite-width neural networks (also GPs).

The goal of this paper is to provide a scalable and online version of MacKay's original proposal, and apply it to modern, larger models used in deep learning.
We propose a scalable Laplace's method to approximate the marginal likelihood and select both the hyperparameters and network architectures. Differentiable hyperparameters can be optimized \emph{during} training in an online fashion (unlike \citet{mackay1995probable}'s original proposal and \citet{khan2019approximate}'s recent approach), and discrete ones can be selected after training.

The method relies on the Generalized Gauss-Newton (\ggn) and the Empirical Fisher (\ef) approximation to the Hessian. The full \ggn and \ef can also be extremely expensive and we reduce the cost by using additional diagonal and block-diagonal approximations~\citep{martens2015optimizing, botev2017practical}.
All these approximations have been used in approximate inference before~\citep{khan2018fast, zhang2018noisy, ritter2018scalable, osawa2019practical, foresee1997gauss, khan2019approximate}, but they have not been applied to model selection and their effectiveness for marginal likelihood estimation has also been unknown.
\changed{Unlike our proposal, the method by  \citet{khan2019approximate} is limited to regression and uses the full but intractable \ggn approximation. \citet{blundell2015weight} also used the variational lower-bound for hyperparameter optimization but found it to give worse results.}
Our main contribution is to show that, even after making these approximations to improve scalability, the estimated marginal likelihoods can faithfully select reasonable models (see \cref{fig:toy_model_selection,fig:teaser_marglik_accuracy}). Our method achieves performance on par or better than cross-validation on a range of regression and classification benchmarks. The best architecture identified by the marginal likelihood aligns with the test performance (see \cref{fig:teaser_marglik_accuracy}). 
\changed{It aligns well with the empirical observation that ResNets often perform better than the standard convolutional networks, which in turn are observed to give better results than their fully-connected counterparts.}
\changed{Overall, our work supports the long-held hypothesis that the marginal likelihood is effective for model selection in deep learning, and shows that a relatively cheap and simple approximation can achieve competitive results.}

\section{Background}%
\label{sec:background}
In this paper, we consider supervised learning tasks with inputs $\vx_n \in \R^D$ and $C$-dimensional vector outputs $\vy_n$ (real or categorical outcomes), and denote the data as the set $\D=\{(\vx_n, \vy_n)\}_{n=1}^N$ of $N$ training-example pairs.

\paragraph{Bayesian models.} We denote by $\vf(\vx, \param)$ the $C$-dimensional real-valued output of a neural network with parameters $\param\in \R^P$, specified by a model \model which typically consists of a network architecture and \emph{hyperparameters}.
A Bayesian model can then be defined using a likelihood and a prior, to get the posterior distribution: $p(\param|\data,\model) \propto p(\data|\param,\model) p(\param|\model)$. We assume that the data examples are sampled \iid~from $p(\vy|\vf(\vx,\param),\model)$.
The normalizing constant of the posterior, also known as the \emph{marginal likelihood}, is given by the following expression:
\begin{equation}
  \label{eq:marglik}
  p(\data|\model) = \textstyle\int \prod_{n=1}^N p(\vy_n|\vf(\vx_n,\param),\model) p(\param|\model) \, \mathrm{d} \param .
\end{equation}
The model \model might consist of the choice of network-architecture (CNN, ResNet, etc.) and hyperparameters of the likelihood and prior, for example, observation noise and prior variances.
Some of these are continuous parameters while others are discrete.
Our goal is to use the marginal likelihood to select such parameters.

\paragraph{Bayesian model comparison.}
The marginal likelihood in \cref{eq:marglik} can be seen as a probability distribution over the space of all datasets (of size $N$) given \changed{the model} $\model$.
The distribution is expected to be wider for complex models, because such models can generate data of more variety than simpler models.
Given the training data $\data$, a model too simple or too complex will therefore be assigned a lower probability $p(\data|\model)$, naturally yielding \emph{Occam's razor} \citep{blumer1987occam, Jefferys1992_occams_razor, rasmussen2001occam}. See Fig.~3.13 in \citet{bishop2006pattern} for an illustration.
A simple method is to pick the model that assigns the highest probability to the training data,
\begin{align}
  \label{eq:model_selection}
  \modelmap &= \textstyle\argmax_\model p(\data \given \model).
\end{align}
This procedure is also called \emph{type-II maximum likelihood} estimation or \emph{empirical Bayes} \citep{mackay2003information}; it is commonly used in the Gaussian process literature to find the hyperparameters of the kernel function \citep{rasmussen2006gaussian}. %
Additionally, we can reapply Bayes rule with the marginal likelihood as the \emph{likelihood} and an additional prior distribution $p(\model)$ over the models \citep{mackay1992practical,mackay2003information}.
\cref{eq:model_selection} can then be seen as a special case of a \emph{maximum a-posteriori} (MAP) estimate with a uniform prior.

\paragraph{Laplace's method.}
Using the marginal likelihood for neural-network model selection was originally proposed by \citet{mackay1992practical}, who used Laplace's method to approximate \cref{eq:marglik}.
The method relies on a local quadratic approximation of $\log p(\param\given\data, \model)$, around a maximum $\paramstar$, resulting in a Gaussian approximation to $p(\param|\data,\model)$, denoted by $q(\param|\data,\model)$, and an approximation to the marginal likelihood, denoted by $q(\data\given\model)$ and shown below,
\begin{align}
  \log p(\data \given \model) &\approx \log q(\data \given \model) \label{eq:lap_marglik}\\
  &:= \log p(\data, \paramstar \given \model) - \half \log \left\vert\tfrac{1}{2\pi}\mH_{\paramstar} \right\vert,  \nonumber%
\end{align}
with Hessian $\mH_\param := -\nabla_{\param \param}^2 \log p(\data, \param \given \model)$. %
However, computing $\mH_\param$, a large matrix of size ${P\times P}$, and its determinant is infeasible in general.
\changed{We refer to the log marginal likelihood as \emph{margLik} and report its value normalized by the number of training examples.}

Using the above objective to select models can also lead to better generalization. %
The log-determinant in \cref{eq:lap_marglik} favors Hessians with small eigenvalues and is thus closely related to low curvature and flatness of a solution's neighborhood.
Such \emph{flat minima} have been argued to generalize better than sharper ones \citep{hochreiter1997flat}, a claim that is supported empirically \citep{keskar2016large, jiang2019fantastic, maddox2019simple} and has also been formalized using PAC-Bayesian generalization bounds \citep{dziugaite2017computing}.
We can therefore expect models selected using the marginal-likelihood to have similar properties.
\changed{The local Gaussian approximation above may not always be sufficient, and the goal of this paper is to investigate its capacity to measure complexity of large neural networks.}

\paragraph{Hessian approximations.} Recently, scalable approximations to the Hessian $\mH_\param$, such as those based on the generalized Gauss-Newton (\ggn)  and empirical Fisher (\ef), have been applied to approximate Bayesian inference~\citep{khan2018fast, zhang2018noisy, ritter2018scalable, osawa2019practical, immer2020improving, kristiadi2020being}, but their application to marginal-likelihood estimation or model selection has not yet been explored, \changed{except by \citet{khan2019approximate} on small regression tasks with a full \ggn}.

The \ggn approximation to the log-joint Hessian is based on the \emph{Jacobian} matrix $\jac \in \R^{C \times P}$ of the network features with entries $\left[\jac\right]_{ci} = \frac{\partial f_c(\vx, \param)}{\partial \theta_i}$, as well as the Hessians of the log-likelihood and the log-prior,
\begin{align*}
    \mLambda(\vy; \vf) &:= - \nabla^2_{\vf \vf}\log\onelikelihoodf, \quad
    \mP_\param := -\nabla_{\param \param}^2 \log p(\param \given \model),
\end{align*}
respectively. The \ggn approximation is then given by
\begin{align}
  \begin{aligned}
    \mH_\param \approx \mH^{\ggn}_\param = \mJ_\param \transpose \mL_\param \mJ_\param + \mP_\param,
    \label{eq:ggn}
  \end{aligned}
\end{align}
where $\mJ_\param$ is an ${NC \times P}$ matrix formed by \emph{stacking} $N$ Jacobians $\jacn$ overall $\vx_n$ in $\data$, and $\mL_\param$ is a $NC\times NC$ block-diagonal matrix with blocks $\mLambda(\vy_n;\vf(\vx_n,\param))$.

The expression for the \ggn approximation in \cref{eq:ggn} is equivalent to the Fisher information matrix~\citep{kunstner2019limitations}; we therefore also consider the \emph{empirical Fisher} (\ef) approximation to the Hessian, which relies on an outer product of the gradients $\mG_\param(\vx) = \nabla_\param \log p(\vy|\vf(\vx,\param),\model) \in \R^P$:
\begin{equation}
  \begin{aligned}
    \mH_\param \approx \mH_\param^{\ef} = \mG_\param \transpose \mG_\param + \mP_\param,
    \label{eq:efisher}
  \end{aligned}
\end{equation}
where $\mG_\param$ is an ${N\times P}$ matrix of stacked gradients $\mG_\param(\vx_n)$ for all $\vx_n\in\data$.
The complexities of the \ggn and \ef are \order{P^2NC+PNC^2} and \order{P^2N}, respectively.
The \ef is therefore \order{C} cheaper to compute as $C$ is typically much smaller than $P$.
For details, see \cref{app:computation}.
Throughout, we use "Laplace-\ggn{}" and "Laplace-\ef{}" to refer to Laplace's method where the Hessian has been approximated by using the \ggn or \ef, respectively.

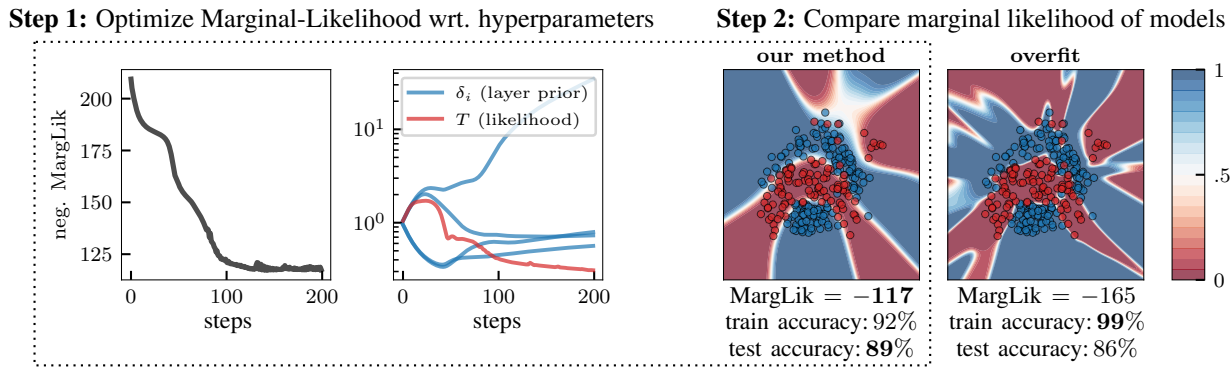
\begin{figure*}[t]
  \centering
  \begin{tikzpicture}
    \node (left){\includegraphics[width=\columnwidth]{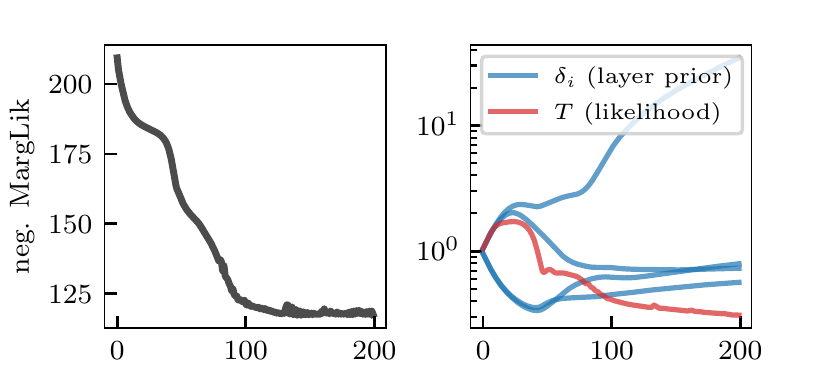}};
    \node[right=-0.5cm of left] (right){\includegraphics[width=\columnwidth]{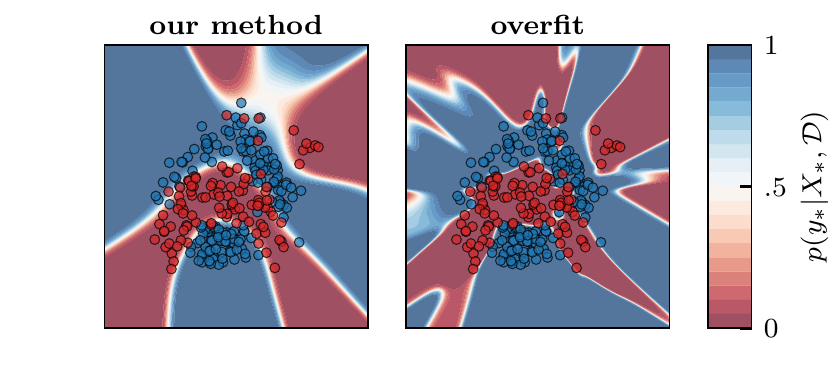}};
    \node[anchor=south] at ($(left.north)+(-0.3,-0.2)$) {\textbf{Step 1:} Optimize Marginal-Likelihood wrt. hyperparameters};
    \node[anchor=south] at ($(right.north)+(+0.2,-0.2)$) {\textbf{Step 2:} Compare marginal likelihood of models};

  \node at ($(left.south)+(-1.68,0.)$) {
      \begin{tikzpicture}
        \node [text width=3cm,align=center] {\small steps};
      \end{tikzpicture}
  };
  \node at ($(left.south)+(1.9,0.)$) {
      \begin{tikzpicture}
        \node [text width=3cm,align=center] {\small steps};
      \end{tikzpicture}
  };

  \node at ($(right.south)+(-1.8,0.0)$) {
      \begin{tikzpicture}
        \node [text width=3cm,align=center] {\small $\text{MargLik}= \mathbf{-117}$ \\ $\text{train accuracy:}\, 92\% $ \\[-0.2em] $ \text{test accuracy:}\, \mathbf{89\%}$};
      \end{tikzpicture}
  };
  \node at ($(right.south)+(1.2,0.0)$) {
      \begin{tikzpicture}
        \node [text width=3cm,align=center] {\small $\text{MargLik}= -165$ \\ $\text{train accuracy:}\, \mathbf{99\%}$ \\[-0.2em] $ \text{test accuracy:}\, 86\%$};
      \end{tikzpicture}
  };
  \draw[black,thick,dotted] ($(left.north west)+(0,-0.2)$)  rectangle ($(right.south east)+(-4.6,-0.55)$);
  \end{tikzpicture}
  \vskip -0.75em
  \caption{
  Proposed method for model selection using the marginal likelihood. In Step 1, we apply our online algorithm (\protect\cref{alg:online_marglik}) to optimize the marginal likelihood estimate (\cref{eq:lap_marglik}) with respect to the differentiable hyperparameters (\textit{here}: prior precision $\delta_i$ per layer and softmax temperature $T$).
  In Step 2, we compare the resulting model (\textit{left}) to an overfitting model (\textit{right}) with higher training accuracy but lower test accuracy; both models have the same architecture. The Laplace-\ggn marginal likelihood estimate $\log q(\data\given\model)$ correctly identifies the model that generalizes better. See \cref{sec:toy_examples} for details.
    }
  \label{fig:toy}
\end{figure*}

\section{Model Selection with Laplace-\ggn and -\ef}%
\label{sec:method}
We propose to use \ggn and \ef approximations for scalable marginal-likelihood estimation based on Laplace's method. 
Our method involves two key steps:
\begin{description}[leftmargin=1.em,rightmargin=0em,itemsep=0.1em,topsep=0pt, partopsep=0pt]
    \item[Step 1.] During training, we update differentiable hyperparameters of \model by using gradients of \cref{eq:lap_marglik}. See line 8 in \cref{alg:online_marglik} and cf. \cref{fig:toy} \textit{(left)} for an example.
    \item[Step 2.] After training, we use the marginal likelihood of the trained model to make discrete choices, e.g., to select the network architecture as shown in \cref{fig:toy_model_selection}, or to choose between several trained models, shown in \cref{fig:toy} \textit{(right)}.
\end{description}
We first describe each of these steps and then give details about efficient Hessian-determinant computations.

\subsection{Step 1: Online model selection during training}
\label{sec:online_model_selection}

Here we alternate between optimizing the network parameters $\param$ and updating the continuous hyperparameters $\model^\partial\subseteq\model$ that appear in the marginal likelihood in a differentiable way.
Such differentiable hyperparameters may include the observation noise of the likelihood, the variances of the Gaussian prior, or the softmax temperature parameter.
The resulting algorithm is outlined in \cref{alg:online_marglik}.

To optimize the network parameters $\param$ we perform regular neural network training on the \textit{maximum a posteriori} (MAP) objective in \cref{eq:map_objective} using stochastic optimizers like \adam \citep{kingma2014adam} or SGD~(cf. line 4 in \cref{alg:online_marglik}),
\begin{align}
  \hspace*{-0.5em}\log p(\data, \param \given \model) &= \sum_{n=1}^N \log \likelihoodn + \log \prior.
  \label{eq:map_objective}
\end{align}
To optimize the continuous hyperparameters $\model^\partial$, we perform gradient ascent on the marginal likelihood estimate (\cref{eq:lap_marglik}). Because of computational considerations further detailed in \cref{sub:computational}, we only do so every $F$ epochs after an initial burn-in period of $B$ epochs and then perform $K$ update steps with step size $\gamma>0$ (cf. lines 6-10 in \cref{alg:online_marglik}):
\begin{equation}
  \model^\partial \, \la \,  \model^\partial + \gamma \nabla_{\model^\partial} \log q(\data \given \model).
  \label{eq:hyperparam_update}
\end{equation}
The marginal likelihood estimate can be updated cheaply for each of the $K$ iterations~(cf.~\cref{app:computation}).
\cref{fig:toy} \textit{(left)} illustrates this optimization of the Gaussian prior-variances for each layer as well as the softmax temperature of the likelihood. We can also use the marginal-likelihood for early stopping \emph{without} validation data.
Note that the parameters $\param_*$ in \cref{eq:lap_marglik} are assumed to be the MAP estimate, however this is not true during training at some $\param$.
We also try another method derived from a local integration in \cref{app:marglik} instead, but empirically this does not yield good results and is more expensive.
We discuss the choice of hyperparameters of \cref{alg:online_marglik} in \cref{sub:computational}.

\begin{algorithm}[t]
   \caption{Marginal likelihood based training}
   \label{alg:online_marglik}
\begin{algorithmic}[1]
  \STATE {\bfseries Input:} dataset $\data$, likelihood $p(\data \given \param, \model)$, prior $p(\param \given \model)$. Initial model $\model$, step size $\gamma$, steps $K$, burn-in epochs $B$, marglik frequency $F$. \ggn or \ef.
  \STATE initialize $\param$ of the neural network
  \FOR{each $epoch$}
    \STATE $\param\, \la \, trainEpoch(\text{objective in \cref{eq:map_objective}})$
    \IF{$epoch > B$ and $epoch \mod F = 0$}
    \STATE compute $\log q(\data \given \model)$~(\cref{eq:lap_marglik}) with $\mH_\param^\ggn$ or $\mH_\param^\ef$
    \FOR{$K$ steps}
    \STATE $\model^\partial \, \la \, \model^\partial + \gamma \nabla_{\model^\partial} \log q( \data \given \model)$
    \STATE update $\model$ and $\log q(\data \given \model)$
    \ENDFOR
   \ENDIF
 \ENDFOR
 \STATE Return marginal likelihood $\log q(\data \given \model)$ and optionally posterior approximation $q(\param \given \data, \model)$.
\end{algorithmic}
\end{algorithm}

\subsection{Step 2: Model selection after training}
To choose between two discrete model alternatives, such as different architectures, we compare their marginal likelihood estimates after training.
This step is a basic hypothesis test, where we compare two models $\model$ and $\model'$ and choose the more likely model given the data according to the likelihood ratio ${p(\data \given \model)}/{p(\data \given \model')}$, which is the most powerful statistical test for this purpose~\citep{neyman1933ix}.
In terms of the marginal likelihood, we only need to choose the model with a higher value~(cf. \cref{fig:toy_model_selection} and \cref{fig:toy} \textit{(right)}).

\subsection{Scalable Laplace approximations}

\paragraph{Efficient determinant computation.}
Scalable marginal likelihood estimation (\cref{eq:lap_marglik}) relies on an efficient computation of the determinant of the \ggn or \ef approximation of the Hessian (\cref{eq:ggn,eq:efisher}).
When $N$ is small, we can use the Woodbury matrix identity to rewrite the determinant of the Hessian (a $P \times P$ matrix) in terms of determinants of matrices whose size only depends on the number of data points $N$ and outputs (e.g., classes) $C$ instead:
\begin{align}
  \hspace*{-1.em}| \mH^\ggn_\param |&= |\mathunderlinewithtext{\mJ_\param \transpose \mL_\param \mJ_\param + \mP_\param}{P\times P}| =|\mathunderlinewithtext{\mJ_\param \mP_\param\inv \mJ_\param\transpose + \mL_\param\inv}{NC\times NC}|  | \mL_\param| | \mP_\param|\nonumber\\[10pt]%
  | \mH^\ef_\param | &= |\mathunderlinewithtext{\mG_\param \transpose \mG_\param + \mP_\param}{P \times P}| = | \mathunderlinewithtext{\mG_\param \mP_\param\inv \mG_\param\transpose + \mI_N}{N\times N} | | \mP_\param |. \label{eq:det_ef}
\end{align}
\vskip0.1em %
The determinants $|\mP_\param|$ (though still $P\times P$) and $|\mL_\param|$ are usually cheap to compute as the prior $\prior$ often factorizes across parameters and $\mL_\param$ is block-diagonal. When neither \order{N^3} nor \order{P^3} are tractable, we consider the following structured \ggn approximations of different sparsities.%

\paragraph{Kronecker-factored Laplace.}
The Kronecker-factored (\kfac) \ggn approximation is based on a block-diagonal approximation to $\mH^\ggn_\param$ and is specified by a Kronecker product per layer~\citep{martens2015optimizing, botev2017practical}.
The \ggn of the $l$-th layer of the neural network is approximated as $[\mJ_\param\transpose \mL_\param \mJ_\param]_l \approx \mQ_l \kron \mW_l$ where $\mQ_l$ is computed from the gradient by backpropagation and $\mW_l$ depends on the input to the $l$-th layer.
$\mW_l$ and $\mQ_l$ are quadratic in the $l$-th layer's input and output size, respectively.
Let $\vq^{(l)} \in \R^{D_l}$ and $\vw^{(l)}\in \R^{D'_l}$ be the eigenvalues of $\mQ_l$ and $\mW_l$, respectively.
If the prior Hessian $\mP_\param$ is isotropic per layer, that is, the $l$-th block-diagonal entry of $\mP_\param$ is given by $p_\theta^{(l)} \mI_l$,
then we can compute the determinant for the Laplace-\ggn efficiently as
\begin{equation}
  | \mH^\ggn_\param | \approx | \mH^\kfac_\param | = \textstyle\prod_l \prod_{ij} \vq^{(l)}_i \vw^{(l)}_j + p_\theta^{(l)}.
  \label{eq:det_kron}
\end{equation}
In contrast to the typical use of Kronecker-factored approximations in optimization~\citep{martens2015optimizing, botev2017practical} and approximate inference~\citep{ritter2018scalable, zhang2018noisy}, we do not use damping which would distort the Laplace-\ggn~(cf. \cref{app:kfac} for discussion and ablation experiment).
Computationally, the Kronecker-factored Laplace-\ggn is cheaper than the full Laplace-\ggn because we only need to decompose matrices that are quadratic in the number of neurons per layer.
This number typically does not exceed a few thousand.

\paragraph{Diagonal~Laplace.} %
The diagonal Laplace approximates the \ggn or \ef by their diagonal, which allows for the cheap computation of the determinants
\begin{equation}
  | \mH^\ggn_\param | \approx \textstyle\prod_p \sqr{\mH^\ggn_\param}_{pp} \, \text{and} \,\, | \mH^\ef_\param | \approx \prod_p \sqr{ \mH^\ef_\param } _{pp}.
\end{equation}
Here, $[\mH]_{pp}$ denotes the diagonal entries of $\mH$.
The computation of the diagonal \ggn still scales with the number of outputs $C$ due to the dependence on the block-diagonal $\mL_\param$.
In contrast, the diagonal \ef is very cheap to compute because it only requires the sum of point-wise squared individual gradients over the training data.
Despite their simplicity \changed{and the hypothesis by \citet{mackay1992practical} that they are ``no good''}, we find that these approximations work well in practice.
\changed{Especially for convolutional neural networks, the diagonal approximations provide a cost-effective alternative to block-diagonal approximations when used in \cref{alg:online_marglik}.}
This is in contrast to the predictive performance of diagonal Laplace approximations, which are often significantly worse  their less sparse counterparts~\citep{ritter2018scalable, immer2020improving}.

\subsection{Computational considerations}%
\label{sub:computational}

The overall runtime of our algorithm depends on the frequency $F$ with which we estimate the marginal likelihood, the number of consecutive hyperparameter update steps $K$, as well as the burn-in period $B$. Furthermore, the cost of each marginal likelihood estimation~(line 6 in \cref{alg:online_marglik}) and marginal likelihood update~(line 9 in \cref{alg:online_marglik}) depends on the specific approximation of the Hessian used. Here, we discuss the most important aspects briefly.
See \cref{app:computation} for more details and discussion, including computation and memory complexities.

Estimation of the marginal likelihood (\cref{eq:lap_marglik}) requires one full pass over the training data: While its first term, the log joint $p(\data,\param\given\model)$ (\cref{eq:map_objective}), decomposes into individual terms per data point and can, therefore, be estimated on minibatches, this is not the case for the log determinant of the Hessian. However, this initial estimation cost amortizes over the $K$ consecutive update steps for the hyperparameters (\cref{alg:online_marglik} lines $7$ to $10$), as the most expensive computations can be reused.
Therefore, it makes sense to perform many update steps ($K$ large) but do so at a lower frequency (estimation frequency $F$ low). %
We typically use $K=100$ update steps in large-scale experiments and perform marginal likelihood estimation every $F=1$ epochs for small-scale experiments or $F = 5$ to $10$ epochs for larger experiments.
\changed{In contrast to \citet{mackay1995probable}, these updates are continuous during training, instead of a single update and re-training which would be prohibitive}.

We use \adam for the gradient updates of the hyperparameters in \cref{eq:hyperparam_update} and line $8$ of \cref{alg:online_marglik}.
We can use a slightly higher learning rate than default ($\gamma$ between $0.01$ and $1$) without divergence because the gradients are not stochastic~\citep{bottou2010large}.
Similar to the approach by \citet{lyle2020bayesian}, our marginal likelihood estimate can be used for early stopping or to save models so that we end up using the model that achieved the best marginal likelihood during training in \cref{alg:online_marglik}.
Overall, we find that the online algorithm is robust to different settings of $K$, $F$, $B$, and $\gamma$, and it is not difficult to monitor convergence of the marginal likelihood and hyperparameters, of which there are fewer than model parameters.
In our experiments, we reliably optimize up to around $400$ hyperparameters in the case of a ResNet.
Typically, our algorithm does not need more epochs to converge than standard training.
For further discussion of the robustness to hyperparameters, see \cref{app:exp:ablation}.

\section{Experiments}
\label{sec:experiments}

We compare our method to model selection by cross-validation and also to manually selected models.
Depending on the problem size, we use different variants of the Laplace-\ggn approximation.
On smaller scale UCI~\citep{ucidata} and toy examples, we use the full \ggn and \ef determinants and the Kronecker-factored approximation.
On larger problems, we use the Kronecker-factored and the diagonal Laplace-\ggn and \ef approximations.

We also compare the posterior predictive obtained by a Laplace-\ggn posterior approximation~\citep{ritter2018scalable, immer2020improving}, defined in \cref{eq:linpred}, to the MAP prediction, defined in \cref{eq:MAPpred},
\begin{align}
    p_\text{dist}(\vy^\ast\given\vx^\ast, \data) & = \tfrac{1}{S}\textstyle\sum_s^S p(\vy^\ast\given \vf_\text{lin}^{\paramstar}(\vx^\ast, \param_s), \model_\ast), \label{eq:linpred}\\
    p_\text{MAP}(\vy^\ast\given\vx^\ast, \data) & = p(\vy^\ast\given \vf(\vx^\ast, \paramstar), \model_\ast). \label{eq:MAPpred}
\end{align}
Here, $\vf_\text{lin}^{\paramstar} (\vx, \param_s) = \vf(\vx,\paramstar) + \mJ_{\paramstar}(\vx) (\param - \paramstar)$ is the linearized neural network at $\paramstar$, and $\param_s \sim q(\param\given \data, \model_\ast)$.

In all experiments, we optimize the precision parameter $\delta_l>0$ of a Gaussian prior $\gauss(0, \delta_l^{-1}\mI_l)$ for the weights and biases of each layer $l$ individually and initialize all $\delta$ to $1$.
In regression experiments, we additionally optimize the Gaussian observation noise.
We use \adam~\citep{kingma2014adam} to optimize the hyperparameters during training with default settings but a higher learning rate between $0.01$ and $1$~(cf.~\cref{sub:computational}).
We optimize the network parameters using \adam except for the ResNet experiments, where we follow common practice and use SGD with momentum of $0.9$.
On the image classification datasets, train for $300$ epochs in total and decay the learning rate by a factor of $0.1$~\citep{he2016residual} after $150, 225$, and $275$ epochs, respectively.

\subsection{Illustrative examples}%
\label{sec:toy_examples}
First, we illustrate our model selection approach on simple regression~(\cref{fig:toy_model_selection}) and classification~(\cref{fig:toy}) examples.
\begin{figure}[t]
    \vskip -1.25em
    \centering
    \includegraphics[width=\columnwidth, trim=0.08cm 0 0.6cm 0, clip]{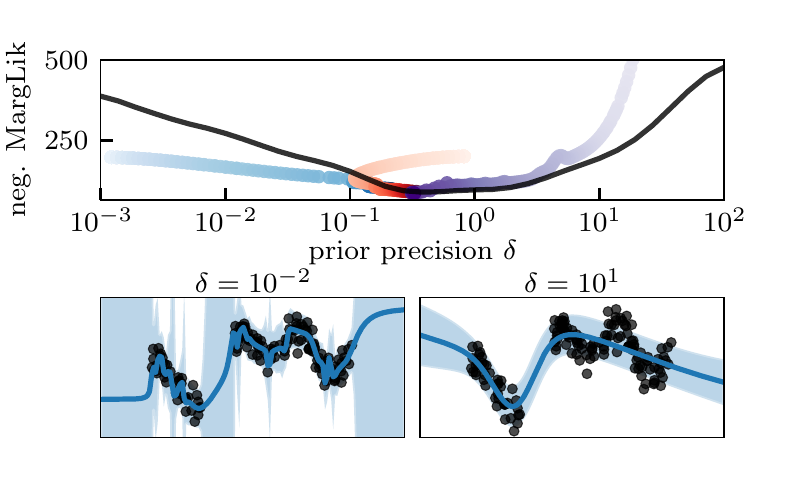}
    \vskip -2.25em
    \caption{
    Selection of the prior precision $\delta$ for a $3$-layer network using the marginal likelihood. \textit{Top:}
    Three optimization traces (init\,
    \protect\tikz[baseline=-.75ex,inner sep=0pt]{
    \protect\node[C1,scale=1.5, opacity=0.2] at (0,0)   {\pgfuseplotmark{*}};
    \protect\node[C1,scale=1.5, opacity=0.4] at (0.15,0) {\pgfuseplotmark{*}};
    \protect\node[C1,scale=1.5, opacity=0.6] at (0.3,0) {\pgfuseplotmark{*}};
    \protect\node[C1,scale=1.5, opacity=0.8] at (0.45,0) {\pgfuseplotmark{*}};
    \protect\node[C1,scale=1.5, opacity=1.0] at (0.6,0) {\pgfuseplotmark{*}};}\,
    final) of the online algorithm compared to a grid-search with fixed $\delta$
    (\protect\tikz[baseline=-0.25ex,inner sep=0pt]{\protect\draw[line width=1.5pt] (0,.05) -- ++(0.55,0);}).
    The online algorithm is initialized with three different prior precisions $\delta_\text{init}\in\{10^{-3}
    (\,\protect\tikz[baseline=-.75ex,inner sep=0pt]{\protect\node[C1,scale=1.5, opacity=0.65] at (0,0) {\pgfuseplotmark{*}};}\,)
    , 1
    (\,\protect\tikz[baseline=-.75ex,inner sep=0pt]{\protect\node[C4,scale=1.5, opacity=0.65] at (0,0) {\pgfuseplotmark{*}};}\,)
    , 10^2
    (\,\protect\tikz[baseline=-.75ex,inner sep=0pt]{\protect\node[C5,scale=1.5, opacity=0.65] at (0,0) {\pgfuseplotmark{*}};}\,)
    \}$ and converges to the optimum as found by the grid-search within the same number of steps.
    \textit{Bottom:} Predictive distributions of an over- and underfitting model, respectively.
    The predictive of the best model is depicted in \cref{fig:toy_model_selection} \textit{(left)}. The marginal likelihood correctly identifies the model that agrees with intuition.
    }
    \vskip-1em
    \label{fig:toy_regression_gridsearch}
\end{figure}

\begin{table*}[h!t]
  \centering
  \footnotesize
  \begin{tabular}{l | RRR | RRR}
  \toprule
    & \multicolumn{3}{c |}{\textbf{cross-validation (CV)}} & \multicolumn{3}{c}{\textbf{MargLik optimization}} \\
    \textbf{Dataset} & \multicolumn{1}{c}{\textbf{MAP}} & \multicolumn{1}{c}{\textbf{VI}} &  \multicolumn{1}{c|}{\textbf{Laplace}} & \multicolumn{1}{c}{\textbf{\ggn}} & \multicolumn{1}{c}{\textbf{\ef}} & \multicolumn{1}{c}{\textbf{\kfac}}  \\ \midrule
    \textbf{boston} & 2.71 {\pms{0.10}} & \mathbf{2.58} \pms{0.03} & \mathbf{2.57 \pms{0.05}} & 2.69 \pms{0.12} & \mathbf{2.62 \pms{0.12}} & 2.70 \pms{0.09} \\
      \textbf{concrete} & 3.17 {\pms{0.04}} & 3.17 \pms{0.01} & \mathbf{3.05 \pms{0.04}} & \mathbf{3.06 \pms{0.05}} & \mathbf{3.07 \pms{0.04}} & 3.13 \pms{0.06}  \\
\textbf{energy} & 1.02 {\pms{0.05}} & 1.42 \pms{0.01} & 0.82 \pms{0.03} & \mathbf{0.55 \pms{0.11} } & 0.73 \pms{0.15}  &  \mathbf{0.53 \pms{0.02} }  \\
\textbf{kin8nm} & -1.09 \pms{0.01} & -0.87 \pms{0.00} & \mathbf{-1.23 \pms{0.01}} & -1.14 \pms{0.01}  & -1.14 \pms{0.01}  & -1.13 \pms{0.01}  \\
\textbf{naval} & -5.75 \pms{0.05} & -3.82 \pms{0.02} & -6.40 \pms{0.06} & \mathbf{-6.93 \pms{0.03}} & \mathbf{-6.92 \pms{0.04}}  &-6.88 \pms{0.05}  \\
\textbf{power} & 2.82 {\pms{0.01}} & 2.86 \pms{0.01} & 2.83 \pms{0.01} & \mathbf{2.78 \pms{0.02} } & \mathbf{2.78 \pms{0.01} } & \mathbf{2.78 \pms{0.02} }  \\
\textbf{wine} & 0.98 {\pms{0.02}} & 0.96 \pms{0.01} & 0.97 \pms{0.02} & \mathbf{0.93 \pms{0.02} } & \mathbf{0.93 \pms{0.01} } & \mathbf{0.94 \pms{0.02} } \\
\textbf{yacht} & 2.30 {\pms{0.02}} & 1.67 \pms{0.01} & \mathbf{1.01 \pms{0.05}} & 5.89 \pms{1.25}  & 2.43 \pms{0.61}  & 1.48 \pms{0.07} \\
  \bottomrule
  \end{tabular}
   \vskip -0.5em
  \caption{Negative test log likelihood (lower is better) on the UCI regression benchmark. Three leftmost columns: cross-validation (CV) using MAP, VI and Laplace predictives; three rightmost columns: our proposed method with \ggn, \ef, and \ggn-\kfac approximation to the marginal likelihood and MAP predictive.
  Even compared to VI and Laplace with CV (second and third column), the MAP predictions obtained with our method perform on par or better except on the kin8nm and yacht datasets.
  The full \ggn approximation performs the best, followed by the \ef and Kronecker Laplace-\ggn approximations.
  The cross-validation results are taken from \citet{foong2019between}.
  Results within one standard error of the best performance are in bold.
  Additional results with diagonal \ggn and \ef as well as Bayesian predictive are in \protect\cref{app:experiments}.}
  \vskip -0.5em
  \label{tab:uci_regression}
\end{table*}

\paragraph{Regression.}
For the regression example we consider the modified Snelson dataset~\citep{snelson2007flexible, khan2019approximate}, see \cref{fig:toy_model_selection}. We run \cref{alg:online_marglik} for two different architectures ($1$ and $3$ hidden layers, $50$ neurons per layer) with a step size of $0.01$ for parameters and hyperparameters and recompute the marginal likelihood after each epoch ($F=1$) and with no burn-in ($B=0$) for $1000$ epochs with $K=1$ hyperparameter updates per step.
We use the Laplace-\ggn with a full \ggn determinant and compute the Bayesian predictive in \cref{eq:linpred} based on the optional Laplace-\ggn approximation.

The larger network ($P=5221$ parameters) fits the data well and does not overfit due to the optimized regularization (\cref{fig:toy_model_selection} \textit{(left)}).
In contrast, the smaller network ($P=151$ parameters) underfits the data due to its limited capacity and explains parts of the sinusoidal signal with a higher predictive uncertainty (\cref{fig:toy_model_selection} \textit{(right)}).
The marginal likelihood identifies the better model.
In \cref{app:exp:illustrative}, we show in addition how the resulting Bayesian predictive decomposes into \emph{epistemic} and \emph{aleatoric} uncertainty. %
Such a decomposition is only possible with optimized hyperparameters and would require a large grid-search without our online method~\citep{khan2019approximate}.

The larger model has enough capacity to overfit on this dataset. To illustrate the marginal likelihood's ability to select the right amount of regularization and to test the stability of our algorithm, we perform a grid search over the prior precision for the larger network and compare the selected hyperparameters.
The online algorithm reliably converges to the same optimum also identified by comparing the marginal likelihood over the grid search after training (optimization traces in \cref{fig:toy_regression_gridsearch} \textit{(top)}). Moreover, the marginal likelihood correctly identifies the value with the best generalization performance (test log likelihood in \cref{app:exp:illustrative}); choosing hyperparameter values that correspond to worse marginal likelihoods results in overfitting or underfitting respectively, see the predictive distributions in \cref{fig:toy_regression_gridsearch} \textit{(bottom)} and compare them to the optimal solution in \cref{fig:toy_model_selection} \textit{(left)}.

\paragraph{Classification.}
For binary classification, we consider a subsampled ($N=265$ data points) version of the synthetic 2\changed{D} banana dataset~\citep{ratsch2001soft}.
We train the same neural network~(1 hidden layer with $50$ neurons) twice: once with our online algorithm for $200$ steps with otherwise same settings as for the regression example above;
and a second time with a weakly regularized prior akin to regular deep learning training.
The optimization trace and the evolution of the hyperparameters (softmax temperature as well as prior precision per layer) for our online algorithm is shown in \cref{fig:toy} \textit{(left)}.

The weakly regularized model overfits and has a large generalization gap with a very high accuracy of almost $100\%$ on the train set but only $86\%$ on the test set (\cref{fig:toy}, Step 2 \textit{(right)}).
In contrast, our online method reduces this gap and increases the test accuracy to $89\%$ by optimizing the hyperparameters during training using the marginal likelihood (\cref{fig:toy}, Step 2 \textit{(left)}).
\changed{The optional Bayesian predictive also profits from the optimized hyperparameters~(\cref{fig:app:toy_classification_bayes}).}
A comparison of the marginal likelihood estimates after training correctly identifies the model with smaller generalization gap and better test performance.

\subsection{UCI regression}%
\label{sec:uci_regression}

We compare our method to cross-validation on eight UCI~\citep{ucidata} regression datasets following \citet{hernandez2015probabilistic} and \citet{foong2019between}.
In this setup, each dataset is split into $90\%$ training and $10\%$ testing data and a neural network with a single hidden layer, 50 neurons, and a \ReLU activation is used.
We use the hand-tuned and cross-validated results reported by \citet{foong2019between} who compare MAP, variational inference, and the Laplace approximation.
They use $10\%$ of the training data for validation and optimize hyperparameters by grid-search.
We run \cref{alg:online_marglik} for $10,000$ epochs until convergence with the standard learning rate of \adam for both hyperparameters and parameters, and set frequency $F=1$, $K=1$ hyperparameter gradient steps, and do not use burn-in.
We compute standard errors over $10$ random splits.

The results in \cref{tab:uci_regression} indicate that our online algorithm overall performs better than cross-validated baselines in terms of the test log likelihood.
Using the simple and efficient MAP predictive, our method outperforms or matches the Bayesian predictions with cross-validated hyperparameters on 6 out of 8 datasets.
Our algorithm performs similarly using the full \ggn or \ef, and the Kronecker-factored \ggn performs almost as good as the full \ggn.
See \cref{app:exp:uci_regression} and \cref{app:exp:uci_classification} for further details and results on UCI classification datasets, including diagonal \ggn and \ef approximations.
\changed{Perhaps surprisingly, the diagonal approximations are effective and also competitive with cross-validation.}
We further find that our method can alleviate the need for the posterior predictive as it achieves the same performance using only the MAP predictive~(performance comparison in \cref{app:exp:uci_regression}).

\subsection{Image classification}%
\label{sec:image_classification}

\begin{table*}[h!t]
  \centering
  \footnotesize
  \begin{tabular}{l l | C C | C C C | C C C}
  \toprule
    & & \multicolumn{2}{c |}{\textbf{cross-validation}} & \multicolumn{6}{c}{\textbf{marginal likelihood optimization}} \\
                     & & & & \multicolumn{3}{c|}{\textbf{\kfac}} & \multicolumn{3}{c}{\textbf{diagonal \ef}} \\
    \textbf{Dataset} & \textbf{Model} & \textbf{accuracy} & \textbf{logLik} & \textbf{accuracy} & \textbf{logLik} & \textbf{MargLik} & \textbf{accuracy} & \textbf{logLik} & \textbf{MargLik} \\ \midrule
    \textbf{MNIST} & \textbf{MLP} & 98.22  & -0.061  & 98.38  & -0.053  & -0.158  & 97.05  & -0.095  & -0.553   \\
                   & \textbf{CNN} & 99.40 & \mathbf{-0.017}  & \mathbf{99.46}  & \mathbf{-0.016}  & \mathbf{-0.064}  & \mathbf{99.45}  & -0.019  & -0.134   \\
    \midrule
    \textbf{FMNIST} & \textbf{MLP} & 88.09  & -0.347  & 89.83  & -0.305  & -0.468  & 85.72  & -0.400  & -0.756   \\
                    & \textbf{CNN} & 91.39  & -0.258  & \mathbf{92.06} & \mathbf{-0.233}  & \mathbf{-0.401}  & 91.69  & \mathbf{-0.233}  & -0.570   \\
    \midrule
    \textbf{CIFAR10} & \textbf{CNN} & 77.41  & -0.680  & 80.46  & -0.644  & -0.967  & 80.17  & -0.600  & -1.359   \\
                     & \textbf{ResNet} & 83.73  & -1.060  & \mathbf{86.11} & -0.595  & \mathbf{-0.717}  & \mathbf{85.82 } & \mathbf{-0.464}  & -0.876   \\
  \bottomrule
  \end{tabular}
  \vskip-0.5em
  \caption{Model selection by marginal likelihood compared with cross-validation on image classification. %
    We report the test accuracy, test log likelihood (logLik), and log marginal likelihood (MargLik). Higher is better for all metrics.
    Generally, models selected by our method perform better than the cross-validated models, in particular on CIFAR-10.
    More importantly, higher accuracies correspond to higher marginal-likelihood.
    The marginal likelihood agrees with the intuition that CNNs are better than MLPs and ResNets are better than CNNs.
    Performance reported over 5 random initializations, best performance per dataset within one standard error in bold.
    Cross-validation results on CNN and MLP are taken from \citet{immer2020improving}.
    Results with standard errors and performance of diagonal \ggn are reported in \cref{app:exp:image_classification}.
  }
  \label{tab:imgclassification}
  \vskip -1em
\end{table*}

\begin{figure*}[h!t]
\centering
  \centerline{\includegraphics[width=\textwidth]{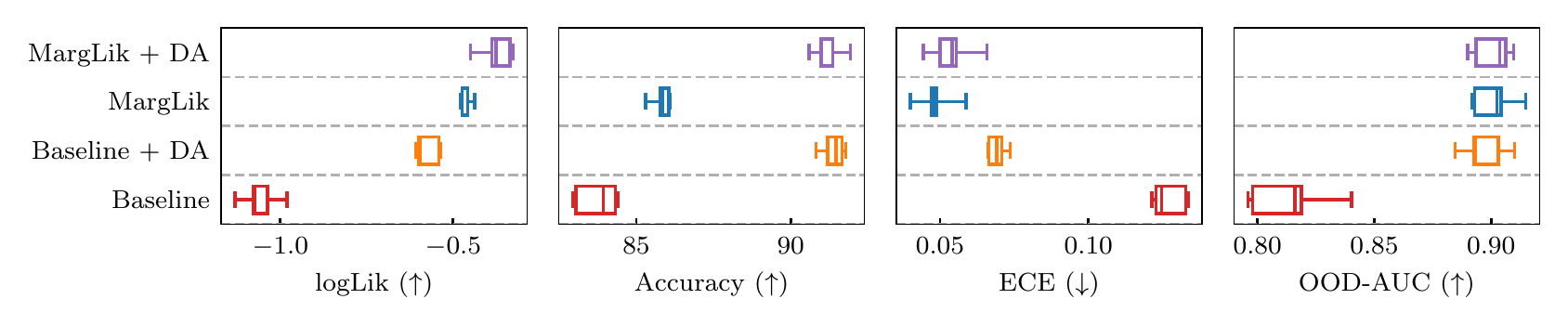}}
\vskip -0.2in
\caption{Comparison to a baseline on the ResNet-20 architecture with and without data augmentation (\emph{DA}) on CIFAR-10.
  We report test log-likelihood (logLik), test accuracy, expected calibration error (ECE), and out-of-distribution (OOD) detection performance in terms of area under the ROC (\emph{OOD-AUC}) using the SVHN validation set as OOD data. Arrows indicate if higher ($\uparrow$) or lower ($\downarrow$) values are better.
  Even without data augmentation, our MargLik-based method achieves a logLik, ECE, and OOD-AUC better than the baseline with data augmentation.
Our approach paired with data augmentation slightly improves on that and matches the accuracy obtained by the baseline with data augmentation.}
\label{fig:resnet_bar}
\end{figure*}

\begin{figure}[t]
\vspace{-1.5em}
\centering
  \centerline{\includegraphics[width=0.9\columnwidth]{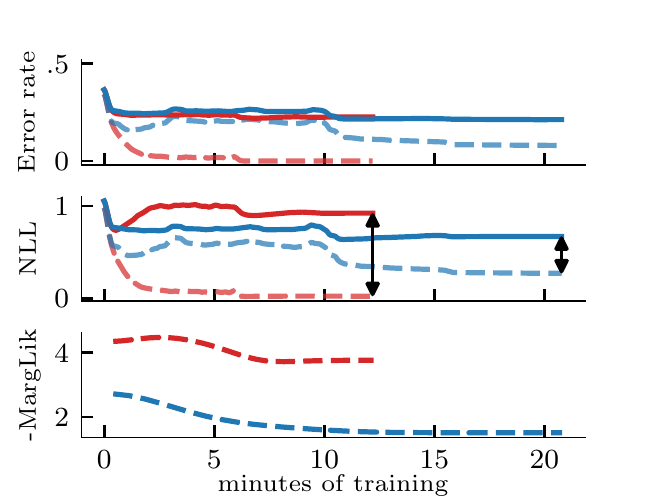}}
  \caption{Train
  (\protect\tikz[baseline=-0.25ex,inner sep=0pt]{\protect\draw[dashed, line width=1.5pt, C1] (0,0) -- ++(0.55,0);\protect\draw[dashed, line width=1.5pt, C4] (0,0.125) -- ++(0.55,0);})
  vs.\ test
  (\protect\tikz[baseline=-0.25ex,inner sep=0pt]{\protect\draw[line width=1.5pt, C1] (0,0) -- ++(0.55,0);\protect\draw[line width=1.5pt, C4] (0,0.125) -- ++(0.55,0);})
  performance of our method
  (\protect\tikz[baseline=-0.25ex,inner sep=0pt]{\protect\draw[line width=1.5pt, C1] (0,0) -- ++(0.55,0);\protect\draw[dashed, line width=1.5pt, C1] (0,0.125) -- ++(0.55,0);})
  compared to a baseline with default weight decay
  (\protect\tikz[baseline=-0.25ex,inner sep=0pt]{\protect\draw[line width=1.5pt, C4] (0,0) -- ++(0.55,0);\protect\draw[dashed, line width=1.5pt, C4] (0,0.125) -- ++(0.55,0);})
  for a CNN on CIFAR-10.
  Online marginal likelihood optimization leads to a smaller generalization gap
  (\protect\tikz[baseline=-0.5ex,inner sep=0pt]{\protect\draw[{Latex[round]}-{Latex[round]},line width=1.pt] (0,0) -- ++(0.55,0);}).
  While the test error rate matches, the baseline has a significantly worse negative log likelihood.
    The Laplace-\ggn marginal likelihood estimate in the bottom indicates that as well.
    The Kronecker-factored Laplace-\ggn approximation increases the training time only by a factor of $2$.
    For cross-validation, this would only suffice to retrain for two different hyperparameters.}
\label{fig:runtime}
\vspace{-1em}
\end{figure}

We benchmark our online model selection against cross-validation on standard image classification datasets~(MNIST, FMNIST, CIFAR-10, CIFAR-100) and use the resulting marginal likelihood estimate to compare architectures.
We compare fully-connected (MLP), convolutional (CNN), and residual (ResNet) networks.
Our algorithm is robust to the selection of its hyperparameters ($F, K, B, \gamma$, see \cref{sec:method} and \cref{app:exp:ablation}).
Here, we use our online model selection step every $F=10$ epochs for $K=100$ hyperparameter steps without burn-in and with step size $\gamma=1$ to keep computational overhead low.
Due to the size of the datasets and models, we use the Kronecker-factored \ggn approximation and the diagonal \ef and \ggn variants.
For the baseline, we use the results and architectures by \citet{immer2020improving} who optimized the prior precision using cross-validation, except for the ResNet on CIFAR-10 where we use the hyperparameters of \citet{zhang2019fixup}.
\changed{For ResNets, we use fixup~\citep{zhang2019fixup} instead of batchnorm because batchnorm conflicts with a Gaussian prior~\citep{zhang2019weight}.}
To estimate standard errors, we run for $5$ different random initializations.

The results in \cref{tab:imgclassification} show that our method improves the performance over baselines on the considered architectures with the largest improvement on CIFAR-10.
We find that the Kronecker-factored version is well-suited for fully-connected networks but the diagonal variants are often sufficient, on CIFAR-10 even slightly better, for convolutional neural networks.
The resulting marginal likelihood estimates of any method suggest that CNNs are better suited for image classification than MLPs, and on CIFAR-10 we further see that ResNets are superior to standard CNNs without residual connections.
In \cref{app:exp:image_classification}, we additionally show the outcome with the diagonal \ggn and list standard errors.

In \cref{fig:resnet_bar}, we show results with additional data augmentation on CIFAR-10 and ResNet architecture and report two additional performance metrics.
\changed{Here, we display the diagonal \ef approximation due to its effectiveness~(performance of Kronecker-factored variant in \cref{app:exp:image_classification}).}
\changed{Additionally to} accuracy and negative log likelihood, we show the expected calibration error (\emph{ECE}, \citet{guo2017calibration}) and out-of-distribution detection performance in terms of area under the receiver-operator curve (\emph{OOD-AUC}).
With and without data augmentation, the models trained with our method achieve better test log-likelihood (\emph{NLL}), expected calibration error (\emph{ECE}), and out-of-distribution detection performance.
Compared to the baseline, the performance can be up to $2\times$ better using our method.
Notably, our method achieves a better NLL, ECE, and similar OOD-AUC \emph{without} data augmentation than the baseline \emph{with} data augmentation.
However, the baseline can mitigate the difference in accuracy by data augmentation and match the performance on our method.
Following \citet{osawa2019practical}, we upweight the likelihood by a factor of $5$ to account for the data augmentation.

In \cref{fig:runtime}, we show that the proposed online model selection can improve over a baseline with weight decay by greatly reducing overfitting and the generalization gap on CIFAR-10.
Compared to a single training run that uses fixed hyperparameters, our online method only takes around twice the time for training when using the Kronecker-factored Laplace-\ggn every $F=10$ epochs.
Cross-validation would only be able to explore two parameter settings in this time.

\begin{figure}
\vspace{-0.5em}
    \centerline{\includegraphics[width=0.495\columnwidth]{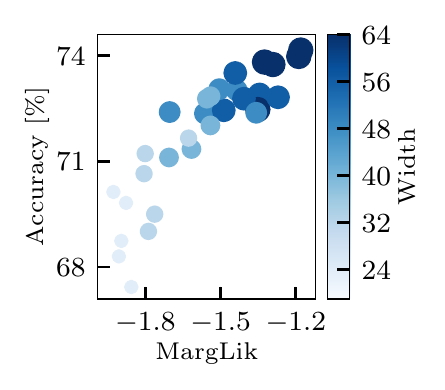}
                \includegraphics[width=0.495\columnwidth]{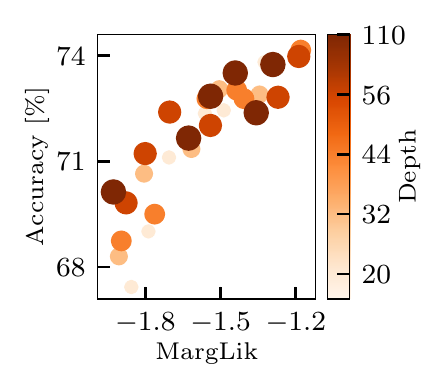}}
                \vspace{-1.em}
    \caption{Wider ResNets give both higher marginal-likelihoods and accuracy (left) while there is no clear trend seen by varying depths (right). Results are on CIFAR-100. Each plot contains $30$ models. Details in \cref{app:exp:model_selection}}
    \label{fig:cifar100_arch}
\vspace{-1em}
\end{figure}

Finally, we conduct a larger-scale study for architecture selection after training (step~2 in~\cref{sec:method}) by applying our algorithm to various architectures and comparing the resulting marginal likelihood estimates.
We use the algorithm with the diagonal \ef approximation due to its efficiency and optimize the prior precision hyperparameter for each layer.\footnote{Similar results can be obtained with the Kronecker-factored variant but at a higher cost.}
After training, we use the final marginal likelihood estimates to rank the different architectures.
On CIFAR-10 and CIFAR-100, we train CNNs and ResNets of varying width (between $2$ and $64$) and depth (between $1$ and $110$).
On FashionMNIST, we compare CNNs and MLPs.
For each data set, we compare around $40$ models.
In \cref{fig:teaser_marglik_accuracy}, each trained model with our method is represented by a marker whose size and color depends on the number of parameters.
On CIFAR-10/100, ResNets achieve higher marginal likelihoods than CNNs despite, in some cases, exponentially more parameters.
In terms of architecture, width tends to improve marginal likelihood more than depth~(see  \cref{fig:cifar100_arch} and \cref{app:exp:model_selection}).
This is in line with improved ResNet architectures that use increased width~\citep{zagoryuko2016wide}.
On both data sets, the rank correlation between test accuracy and marginal likelihood is $97\%$ in terms of Spearman's $\rho$.
Spearman's $\rho$ measures the correlation of the model rankings that test accuracy and marginal likelihood induce~\citep{kendall1948rank}.
CNNs achieve consistently higher marginal likelihoods than MLPs on FMNIST.
Especially at similar performance, CNNs achieve a higher marginal likelihood than MLPs, potentially due to the reduced model complexity.
On FMNIST, the rank correlation is $69\%$.
The correlation is likely lower because the marginal likelihood often selects significantly smaller instead of slightly better performing models (see top right in \cref{fig:teaser_marglik_accuracy}).
For more detailed results and the architectures, see \cref{app:exp:model_selection}.

\section{Conclusion}%
\label{sec:conclusion}
In this paper, we present the marginal likelihood as a viable tool for model selection in deep learning.
Even after making approximations for scalability, the marginal likelihood faithfully reflects the model quality based on the training data alone.
Estimation requires only a mild increase in computation, but allows us to compare models and even select hyperparameters during training.
Ultimately, the marginal likelihood and similar methods make use of the neighborhood of a solution to improve their robustness. %
We believe this is important to design models for applications where little is known about deployment and future usage.
Removing dependence on validation data can widen the impact of deep learning and open up new application areas.

Many interesting research directions remain open for exploration.
It is important to explore ways to further improve Hessian approximations and computation, and consider other approximate inference methods. %
Online model selection for discrete hyperparameters is another interesting direction.
Finally, extending this work beyond supervised learning, e.g., to bandits, Bayesian optimization, active learning, or continual learning, is an important avenue for future research.
We hope that this work encourages others to pursue new ideas based on Bayesian model selection.

\section*{Acknowledgements}
A.~I. acknowledges funding by the Max Planck ETH Center for Learning Systems (CLS).
M.~B. acknowledges funding by the Max Planck Society, the EPSRC, and a Qualcomm European Scholarship in Technology.
V.~F. acknowledges funding by the Swiss Data Science Center.

\bibliography{bibliography.bib}

\begin{thebibliography}{54}
\providecommand{\natexlab}[1]{#1}
\providecommand{\url}[1]{\texttt{#1}}
\expandafter\ifx\csname urlstyle\endcsname\relax
  \providecommand{\doi}[1]{doi: #1}\else
  \providecommand{\doi}{doi: \begingroup \urlstyle{rm}\Url}\fi

\bibitem[Bishop(2006)]{bishop2006pattern}
Bishop, C.~M.
\newblock \emph{Pattern recognition and machine learning}.
\newblock Information Science and Statistics. Springer, 2006.

\bibitem[Blumer et~al.(1987)Blumer, Ehrenfeucht, Haussler, and
  Warmuth]{blumer1987occam}
Blumer, A., Ehrenfeucht, A., Haussler, D., and Warmuth, M.~K.
\newblock Occam's razor.
\newblock \emph{Information processing letters}, 24\penalty0 (6):\penalty0
  377--380, 1987.

\bibitem[Blundell et~al.(2015)Blundell, Cornebise, Kavukcuoglu, and
  Wierstra]{blundell2015weight}
Blundell, C., Cornebise, J., Kavukcuoglu, K., and Wierstra, D.
\newblock Weight uncertainty in neural networks.
\newblock In \emph{Proceedings of the 32nd International Conference on Machine
  Learning}, pp.\  1613--1622, 2015.

\bibitem[Botev et~al.(2017)Botev, Ritter, and Barber]{botev2017practical}
Botev, A., Ritter, H., and Barber, D.
\newblock Practical {G}auss-{N}ewton optimisation for deep learning.
\newblock In \emph{International Conference on Machine Learning}, International
  Convention Centre, Sydney, Australia, 2017. PMLR.

\bibitem[Bottou(2010)]{bottou2010large}
Bottou, L.
\newblock Large-scale machine learning with stochastic gradient descent.
\newblock In \emph{Proceedings of COMPSTAT'2010}, pp.\  177--186. Springer,
  2010.

\bibitem[Buntine \& Weigend(1991)Buntine and Weigend]{buntine1991bayesian}
Buntine, W.~L. and Weigend, A.~S.
\newblock Bayesian back-propagation.
\newblock \emph{Complex systems}, 5\penalty0 (6):\penalty0 603--643, 1991.

\bibitem[Damianou \& Lawrence(2013)Damianou and Lawrence]{deepgp2013}
Damianou, A. and Lawrence, N.~D.
\newblock Deep {G}aussian processes.
\newblock In \emph{Proceedings of the Sixteenth International Conference on
  Artificial Intelligence and Statistics}. PMLR, 2013.

\bibitem[Dangel et~al.(2019)Dangel, Kunstner, and Hennig]{dangel2019backpack}
Dangel, F., Kunstner, F., and Hennig, P.
\newblock Backpack: Packing more into backprop.
\newblock In \emph{Proceedings of 7th International Conference on Learning
  Representations}, 2019.

\bibitem[Dua \& Graff(2017)Dua and Graff]{ucidata}
Dua, D. and Graff, C.
\newblock {UCI} machine learning repository, 2017.
\newblock URL \url{http://archive.ics.uci.edu/ml}.

\bibitem[Dutordoir et~al.(2020)Dutordoir, van~der Wilk, Artemev, and
  Hensman]{deep_gp_image_classification_2020}
Dutordoir, V., van~der Wilk, M., Artemev, A., and Hensman, J.
\newblock Bayesian image classification with deep convolutional gaussian
  processes.
\newblock In \emph{Proceedings of the Twenty Third International Conference on
  Artificial Intelligence and Statistics}, 2020.

\bibitem[Dziugaite \& Roy(2017)Dziugaite and Roy]{dziugaite2017computing}
Dziugaite, G.~K. and Roy, D.~M.
\newblock Computing nonvacuous generalization bounds for deep (stochastic)
  neural networks with many more parameters than training data.
\newblock \emph{arXiv preprint arXiv:1703.11008}, 2017.

\bibitem[Fong \& Holmes(2020)Fong and Holmes]{fong2020marginal}
Fong, E. and Holmes, C.
\newblock On the marginal likelihood and cross-validation.
\newblock \emph{Biometrika}, 107\penalty0 (2):\penalty0 489--496, 2020.

\bibitem[Foong et~al.(2019)Foong, Li, Hern{\'a}ndez-Lobato, and
  Turner]{foong2019between}
Foong, A.~Y., Li, Y., Hern{\'a}ndez-Lobato, J.~M., and Turner, R.~E.
\newblock 'in-between'uncertainty in bayesian neural networks.
\newblock \emph{arXiv preprint arXiv:1906.11537}, 2019.

\bibitem[Foresee \& Hagan(1997)Foresee and Hagan]{foresee1997gauss}
Foresee, F.~D. and Hagan, M.~T.
\newblock Gauss-newton approximation to bayesian learning.
\newblock In \emph{International Conference on Neural Networks (ICNN'97)},
  volume~3, pp.\  1930--1935. IEEE, 1997.

\bibitem[Guo et~al.(2017)Guo, Pleiss, Sun, and Weinberger]{guo2017calibration}
Guo, C., Pleiss, G., Sun, Y., and Weinberger, K.~Q.
\newblock On calibration of modern neural networks.
\newblock In \emph{International Conference on Machine Learning}, pp.\
  1321--1330. PMLR, 2017.

\bibitem[Harville(1998)]{harville1998matrix}
Harville, D.~A.
\newblock Matrix algebra from a statistician's perspective, 1998.

\bibitem[He et~al.(2016)He, Zhang, Ren, and Sun]{he2016residual}
He, K., Zhang, X., Ren, S., and Sun, J.
\newblock Deep residual learning for image recognition.
\newblock In \emph{2016 {IEEE} Conference on Computer Vision and Pattern
  Recognition, {CVPR} 2016, Las Vegas, NV, USA, June 27-30, 2016}, pp.\
  770--778. {IEEE} Computer Society, 2016.

\bibitem[Hern{\'a}ndez-Lobato \& Adams(2015)Hern{\'a}ndez-Lobato and
  Adams]{hernandez2015probabilistic}
Hern{\'a}ndez-Lobato, J.~M. and Adams, R.
\newblock Probabilistic backpropagation for scalable learning of bayesian
  neural networks.
\newblock In \emph{International Conference on Machine Learning}, pp.\
  1861--1869. PMLR, 2015.

\bibitem[Hochreiter \& Schmidhuber(1997)Hochreiter and
  Schmidhuber]{hochreiter1997flat}
Hochreiter, S. and Schmidhuber, J.
\newblock Flat minima.
\newblock \emph{Neural Computation}, 9\penalty0 (1):\penalty0 1--42, 1997.

\bibitem[Immer et~al.(2021)Immer, Korzepa, and Bauer]{immer2020improving}
Immer, A., Korzepa, M., and Bauer, M.
\newblock Improving predictions of bayesian neural nets via local
  linearization.
\newblock In \emph{Proceedings of The 24th International Conference on
  Artificial Intelligence and Statistics}, pp.\  703--711, 2021.

\bibitem[Jefferys \& Berger(1992)Jefferys and
  Berger]{Jefferys1992_occams_razor}
Jefferys, W.~H. and Berger, J.~O.
\newblock Ockham's razor and bayesian analysis.
\newblock \emph{American Scientist}, 80\penalty0 (1):\penalty0 64--72, 1992.
\newblock ISSN 00030996.

\bibitem[Jiang et~al.(2019)Jiang, Neyshabur, Mobahi, Krishnan, and
  Bengio]{jiang2019fantastic}
Jiang, Y., Neyshabur, B., Mobahi, H., Krishnan, D., and Bengio, S.
\newblock Fantastic generalization measures and where to find them.
\newblock In \emph{International Conference on Learning Representations}, 2019.

\bibitem[Kendall(1948)]{kendall1948rank}
Kendall, M.~G.
\newblock Rank correlation methods.
\newblock 1948.

\bibitem[Keskar et~al.(2016)Keskar, Mudigere, Nocedal, Smelyanskiy, and
  Tang]{keskar2016large}
Keskar, N.~S., Mudigere, D., Nocedal, J., Smelyanskiy, M., and Tang, P. T.~P.
\newblock On large-batch training for deep learning: Generalization gap and
  sharp minima.
\newblock \emph{arXiv preprint arXiv:1609.04836}, 2016.

\bibitem[Khan et~al.(2018)Khan, Nielsen, Tangkaratt, Lin, Gal, and
  Srivastava]{khan2018fast}
Khan, M., Nielsen, D., Tangkaratt, V., Lin, W., Gal, Y., and Srivastava, A.
\newblock Fast and scalable bayesian deep learning by weight-perturbation in
  adam.
\newblock In \emph{International Conference on Machine Learning}, pp.\
  2611--2620, 2018.

\bibitem[Khan et~al.(2019)Khan, Immer, Abedi, and Korzepa]{khan2019approximate}
Khan, M. E.~E., Immer, A., Abedi, E., and Korzepa, M.
\newblock Approximate inference turns deep networks into gaussian processes.
\newblock In \emph{Advances in Neural Information Processing Systems}, pp.\
  3088--3098, 2019.

\bibitem[Kingma \& Ba(2015)Kingma and Ba]{kingma2014adam}
Kingma, D.~P. and Ba, J.
\newblock Adam: A method for stochastic optimization.
\newblock In \emph{International Conference on Learning Representations}, 2015.

\bibitem[Kristiadi et~al.(2020)Kristiadi, Hein, and Hennig]{kristiadi2020being}
Kristiadi, A., Hein, M., and Hennig, P.
\newblock Being bayesian, even just a bit, fixes overconfidence in relu
  networks.
\newblock In \emph{International Conference on Machine Learning}, pp.\
  5436--5446. PMLR, 2020.

\bibitem[Kunstner et~al.(2019)Kunstner, Hennig, and
  Balles]{kunstner2019limitations}
Kunstner, F., Hennig, P., and Balles, L.
\newblock Limitations of the empirical fisher approximation for natural
  gradient descent.
\newblock In \emph{Advances in Neural Information Processing Systems}, pp.\
  4158--4169, 2019.

\bibitem[Lakshminarayanan et~al.(2017)Lakshminarayanan, Pritzel, and
  Blundell]{lakshminarayanan2017simple}
Lakshminarayanan, B., Pritzel, A., and Blundell, C.
\newblock Simple and scalable predictive uncertainty estimation using deep
  ensembles.
\newblock In \emph{Advances in neural information processing systems}, pp.\
  6402--6413, 2017.

\bibitem[Llorente et~al.(2020)Llorente, Martino, Delgado, and
  Lopez-Santiago]{llorente2020marginal}
Llorente, F., Martino, L., Delgado, D., and Lopez-Santiago, J.
\newblock Marginal likelihood computation for model selection and hypothesis
  testing: an extensive review.
\newblock \emph{arXiv preprint arXiv:2005.08334}, 2020.

\bibitem[Lyle et~al.(2020)Lyle, Schut, Ru, Gal, and van~der
  Wilk]{lyle2020bayesian}
Lyle, C., Schut, L., Ru, R., Gal, Y., and van~der Wilk, M.
\newblock A bayesian perspective on training speed and model selection.
\newblock \emph{Advances in Neural Information Processing Systems}, 33, 2020.

\bibitem[MacKay(1992)]{mackay1992practical}
MacKay, D.~J.
\newblock A practical bayesian framework for backpropagation networks.
\newblock \emph{Neural computation}, 4\penalty0 (3):\penalty0 448--472, 1992.

\bibitem[MacKay(1995)]{mackay1995probable}
MacKay, D.~J.
\newblock Probable networks and plausible predictions—a review of practical
  bayesian methods for supervised neural networks.
\newblock \emph{Network: computation in neural systems}, 6\penalty0
  (3):\penalty0 469--505, 1995.

\bibitem[MacKay(2003)]{mackay2003information}
MacKay, D.~J.
\newblock \emph{Information theory, inference and learning algorithms}.
\newblock Cambridge university press, 2003.

\bibitem[Maddox et~al.(2019)Maddox, Izmailov, Garipov, Vetrov, and
  Wilson]{maddox2019simple}
Maddox, W.~J., Izmailov, P., Garipov, T., Vetrov, D.~P., and Wilson, A.~G.
\newblock A simple baseline for bayesian uncertainty in deep learning.
\newblock In \emph{Advances in Neural Information Processing Systems}, pp.\
  13132--13143, 2019.

\bibitem[Martens \& Grosse(2015)Martens and Grosse]{martens2015optimizing}
Martens, J. and Grosse, R.
\newblock Optimizing neural networks with kronecker-factored approximate
  curvature.
\newblock In \emph{International conference on machine learning}, pp.\
  2408--2417, 2015.

\bibitem[Mascarenhas(2014)]{mascarenhas2014divergence}
Mascarenhas, W.~F.
\newblock The divergence of the bfgs and gauss newton methods.
\newblock \emph{Mathematical Programming}, 147\penalty0 (1):\penalty0 253--276,
  2014.

\bibitem[Neal(1995)]{neal1995bayesian}
Neal, R.~M.
\newblock \emph{Bayesian Learning for Neural Networks}.
\newblock PhD thesis, University of Toronto, 1995.

\bibitem[Neyman \& Pearson(1933)Neyman and Pearson]{neyman1933ix}
Neyman, J. and Pearson, E.~S.
\newblock Ix. on the problem of the most efficient tests of statistical
  hypotheses.
\newblock \emph{Philosophical Transactions of the Royal Society of London.
  Series A, Containing Papers of a Mathematical or Physical Character},
  231\penalty0 (694-706):\penalty0 289--337, 1933.

\bibitem[Osawa et~al.(2019)Osawa, Swaroop, Khan, Jain, Eschenhagen, Turner, and
  Yokota]{osawa2019practical}
Osawa, K., Swaroop, S., Khan, M. E.~E., Jain, A., Eschenhagen, R., Turner,
  R.~E., and Yokota, R.
\newblock Practical deep learning with bayesian principles.
\newblock In \emph{Advances in Neural Information Processing Systems}, pp.\
  4289--4301, 2019.

\bibitem[Rasmussen \& Ghahramani(2001)Rasmussen and
  Ghahramani]{rasmussen2001occam}
Rasmussen, C.~E. and Ghahramani, Z.
\newblock Occam's razor.
\newblock In \emph{Advances in neural information processing systems}, pp.\
  294--300, 2001.

\bibitem[Rasmussen \& Williams(2006)Rasmussen and
  Williams]{rasmussen2006gaussian}
Rasmussen, C.~E. and Williams, C.~K.
\newblock \emph{Gaussian processes for machine learning}.
\newblock MIT press Cambridge, MA, 2006.

\bibitem[R{\"a}tsch et~al.(2001)R{\"a}tsch, Onoda, and
  M{\"u}ller]{ratsch2001soft}
R{\"a}tsch, G., Onoda, T., and M{\"u}ller, K.-R.
\newblock Soft margins for adaboost.
\newblock \emph{Machine learning}, 42\penalty0 (3):\penalty0 287--320, 2001.

\bibitem[Ritter et~al.(2018)Ritter, Botev, and Barber]{ritter2018scalable}
Ritter, H., Botev, A., and Barber, D.
\newblock A scalable laplace approximation for neural networks.
\newblock In \emph{International Conference on Learning Representations}, 2018.

\bibitem[Robbins(1955)]{robbins1955empirical}
Robbins, H.
\newblock \emph{An empirical Bayes approach to statistics}.
\newblock Office of Scientific Research, US Air Force, 1955.

\bibitem[Schneider et~al.(2018)Schneider, Balles, and
  Hennig]{schneider2018deepobs}
Schneider, F., Balles, L., and Hennig, P.
\newblock Deepobs: A deep learning optimizer benchmark suite.
\newblock In \emph{International Conference on Learning Representations}, 2018.

\bibitem[Snelson(2007)]{snelson2007flexible}
Snelson, E.~L.
\newblock \emph{Flexible and efficient Gaussian process models for machine
  learning}.
\newblock PhD thesis, UCL (University College London), 2007.

\bibitem[van~der Wilk et~al.(2018)van~der Wilk, Bauer, John, and
  Hensman]{van2018learning}
van~der Wilk, M., Bauer, M., John, S., and Hensman, J.
\newblock Learning invariances using the marginal likelihood.
\newblock In \emph{Advances in Neural Information Processing Systems}, pp.\
  9938--9948, 2018.

\bibitem[Wenzel et~al.(2020)Wenzel, Roth, Veeling, {\'S}wiatkowski, Tran,
  Mandt, Snoek, Salimans, Jenatton, and Nowozin]{wenzel2020good}
Wenzel, F., Roth, K., Veeling, B.~S., {\'S}wiatkowski, J., Tran, L., Mandt, S.,
  Snoek, J., Salimans, T., Jenatton, R., and Nowozin, S.
\newblock How good is the bayes posterior in deep neural networks really?
\newblock In \emph{International Conference on Machine Learning}, 2020.

\bibitem[Zagoruyko \& Komodakis(2016)Zagoruyko and
  Komodakis]{zagoryuko2016wide}
Zagoruyko, S. and Komodakis, N.
\newblock Wide residual networks.
\newblock In \emph{{BMVC}}. {BMVA} Press, 2016.

\bibitem[Zhang et~al.(2018)Zhang, Sun, Duvenaud, and Grosse]{zhang2018noisy}
Zhang, G., Sun, S., Duvenaud, D., and Grosse, R.
\newblock Noisy natural gradient as variational inference.
\newblock In \emph{International Conference on Machine Learning}, pp.\
  5852--5861, 2018.

\bibitem[Zhang et~al.(2019{\natexlab{a}})Zhang, Wang, Xu, and
  Grosse]{zhang2019weight}
Zhang, G., Wang, C., Xu, B., and Grosse, R.~B.
\newblock Three mechanisms of weight decay regularization.
\newblock In \emph{{ICLR} (Poster)}. OpenReview.net, 2019{\natexlab{a}}.

\bibitem[Zhang et~al.(2019{\natexlab{b}})Zhang, Dauphin, and
  Ma]{zhang2019fixup}
Zhang, H., Dauphin, Y.~N., and Ma, T.
\newblock Fixup initialization: Residual learning without normalization.
\newblock In \emph{International Conference on Learning Representations},
  2019{\natexlab{b}}.

\end{thebibliography}
\bibliographystyle{icml2021}

\FloatBarrier

\clearpage
\appendix
\onecolumn
\renewcommand\thefigure{\thesection\arabic{figure}}
\setcounter{figure}{0}
\renewcommand\theequation{\thesection\arabic{equation}}
\setcounter{equation}{0}
\renewcommand\thetable{\thesection\arabic{table}}
\setcounter{table}{0}

\linewidth\hsize \toptitlebar {\centering
{\Large\bfseries Scalable Marginal Likelihood Estimation for Model Selection in Deep Learning\\ (Appendix) \par}}
 \bottomtitlebar

\setcounter{page}{1}
\counterwithin{figure}{section}
\counterwithin{equation}{section}

\section{Derivations and additional details}
\subsection{Laplace approximation to the marginal likelihood}%
\label{app:marglik}
We briefly derive the Laplace approximation to the marginal likelihood and discuss the additional correction term that would arise in online estimation.
We derive the local Laplace approximation at an arbitrary point $\paramstar$ similar to the standard Laplace approximation by a local quadratic approximation.
Recall the Hessian $\mH_\param = -\nabla_{\theta \theta}^2 \log p(\data, \theta \given \model)$ and define the gradient $\vg_\param = -\nabla_\theta \log p(\data, \theta \given \model)$ of the MAP objective in \cref{eq:map_objective}.
A second-order Taylor approximation around $\paramstar$ to the log joint distribution is then given by
\begin{equation}
  \log p(\data, \param \given \model) \approx \log p(\data, \paramstar \given \model) - (\param - \paramstar)\transpose \vg_\paramstar - \half (\param - \paramstar)\transpose \mH_\paramstar (\param - \paramstar).
\end{equation}

In the standard Laplace approximation, the second term disappears due to $\vg_\paramstar = \vzero$.
We compute the Laplace approximation to the marginal likelihood by solving the integral in \cref{eq:marglik} using the above approximation to the log joint:
\begin{align}
  p(\data \given \model) &= \int p(\data, \param \given \model) \, \mathrm{d} \param \nonumber \\
                         &\approx \int \exp \sqr{\log p(\data, \paramstar \given \model) - (\param - \paramstar)\transpose \vg_\paramstar - \half (\param - \paramstar)\transpose \mH_\paramstar (\param - \paramstar)} \, \mathrm{d} \param \nonumber  \\
                         &= p(\data, \paramstar \given \model) \int \exp \sqr{-(\param - \paramstar)\transpose \vg_\paramstar - \half (\param - \paramstar)\transpose \mH_\paramstar (\param - \paramstar)} \, \mathrm{d} \param \nonumber \\
                         &= p(\data, \paramstar \given \model) \int \exp \sqr{-\half (\param - \paramstar + \mH_\paramstar\inv \vg_\paramstar)\transpose \mH_\paramstar (\param - \paramstar + \mH_\paramstar\inv \vg_\paramstar) + \half \vg_\paramstar\transpose \mH_\paramstar\inv \vg_\paramstar} \, \mathrm{d} \param \nonumber \\
                         &= p(\data, \paramstar \given \model) \exp(\half \vg_\paramstar\transpose \mH_\paramstar\inv \vg_\paramstar) \int \exp \sqr{-\half (\param - \paramstar + \mH_\paramstar\inv \vg_\paramstar)\transpose \mH_\paramstar (\param - \paramstar + \mH_\paramstar\inv \vg_\paramstar)} \, \mathrm{d} \param \nonumber \\
                         &= p(\data, \paramstar \given \model) \exp(\half \vg_\paramstar\transpose \mH_\paramstar\inv \vg_\paramstar) (2\pi)^\frac{P}{2} |\mH_\paramstar|^{-\half}. \nonumber
\end{align}
To obtain this result, we first complete the square inside the exponent:
\begin{align}
  - (\param - \paramstar)\transpose \vg_\paramstar - \half (\param - \paramstar)\transpose \mH_\paramstar (\param - \paramstar)
      &= - \sqr{(\param - \paramstar)\transpose \vg_\paramstar + \half (\param - \paramstar)\transpose \mH_\paramstar (\param - \paramstar)}  \label{eq:app:correction}\\
      &= - \sqr{\half (\param - \paramstar + \mH_\paramstar\inv \vg_\paramstar)\transpose \mH_\paramstar (\param - \paramstar + \mH_\paramstar\inv \vg_\paramstar) - \half \vg_\paramstar\transpose \mH_\paramstar\inv \vg_\paramstar}. \nonumber
\end{align}
The last step is to solve the Gaussian integral which gives the remaining factors $(2\pi)^\frac{P}{2} |\mH_\paramstar|^{-\half}$.
Taking the $\log$, we obtain the log marginal likelihood approximation presented in \cref{eq:lap_marglik} with an additional correction term $\frac12 \vg_\paramstar\transpose \mH_\paramstar\inv \vg_\paramstar$.
We empirically found that the correction term does not help during the online algorithm.%
For example see \cref{fig:runtime}, where the marginal likelihood estimation remains stable during training without a correction term.
However, the correction term could be computed as efficiently as the determinants we require for our algorithm.
We believe the reason for its worse performance is due to an undesired change of the neural network parameter $\paramstar$ to $\paramstar + \mH_\paramstar\inv \vg_\paramstar$.
The parameter change is due to the integration over $\param$ of \cref{eq:app:correction}, which yields a local Laplace approximation \emph{to the posterior} with mean $\vmu = \paramstar + \mH_\paramstar\inv \vg_\paramstar$.
This parameter update resembles a \emph{full} step of Newton's method  and does typically lead to divergence~\citep{mascarenhas2014divergence}, except, for example, in linear least squares.
We therefore ignore the correction term and treat the current parameter as $\paramstar$.
It is common to assume the gradient is zero, i.e., $\vg_\paramstar=\vzero$, when using the Laplace approximation in deep learning~\citep{ritter2018scalable, kristiadi2020being}.
We follow this simplification with our online approximation.

\subsection{Efficient determinant computation}%
\label{app:kernelized}
We can make the computation of the log determinant more efficient using Woodbury's identity and the same idea applies to the correction term in the local Laplace approximation.
The local Laplace approximation relies on terms that involve expensive computations with the Hessian $\mH_\param$ which we approximate either by the \ggn or \ef.
That is, we have either $\mH \approx \mJ_\param\transpose \mL_\param \mJ_\param + \mP_\param$ (\ggn) or $\mH \approx \mG_\param\transpose \mG_\param + \mP_\param$ (\ef).
The results for the determinant are already presented in \cref{eq:det_ef} and a proof can be found in the literature, e.g., see Theorem 18.1.1 by \citet{harville1998matrix}.
For the correction term, we can similarly shift the computational complexity from the number of parameters to the number of data points.
This can be achieved by standard application of Woodbury's matrix identity;
see for example Theorem 18.2.8 by \citet{harville1998matrix}.

\subsection{Kronecker-factored Laplace-\ggn without damping}%
\label{app:kfac}
Kronecker-factored Laplace typically uses \emph{damping} to combine a Gaussian prior with the Hessian approximation~\citep{ritter2018scalable}.
When using damping, the log likelihood Hessian and log prior Hessian are not added but also multiplied, which distorts the Laplace approximation.
In particular because we optimize the prior parameters, such a distortion is likely to be problematic.
We will first discuss damping and how we can compute the log determinant efficiently without damping, before presenting ablation results on image classification which suggest that damping should be avoided in this context.

Kronecker-factored Laplace approximates the \ggn using two Kronecker factors, i.e., we have $[\mJ_\param\transpose \mL_\param \mJ_\param]_l \approx \mQ_l \kron \mW_l$ where $l$ denotes the index of a single layer.
$\mQ_l$ is quadratic in the number of neurons of the $l$-th layer while $\mW_l$ is quadratic in the number of neurons of the previous layer~\citep{botev2017practical}.
We assume an isotropic prior for the corresponding layer with $[\mP_\param]_l = p_\theta^{(l)} \mI_l$.
Then, the Laplace approximation with damping yields
\begin{equation}
  [\mH_\param^\ggn]_l \approx \mQ_l \kron \mW_l + p_\theta^{(l)} \mI_l \approx (\mQ_l + \sqrt{p_\theta^{(l)}} \mI) \kron (\mW_l + \sqrt{p_\theta^{(l)}}\mI),
\end{equation}
where $\mI$ are of matching size.
Clearly, the prior $p_\theta^{(l)}$ is now multiplied by the Kronecker factors and the update is not purely additive anymore.
However, it is equally efficient to compute the determinant of the Kronecker approximation without damping.
We use the eigendecomposition of $\mQ_l \kron \mW_l$ which can be written as $\mM_l (\diag(\vq^{(l)}) \kron \diag(\vw^{(l)})) \mM_l\transpose$, where $\mM_l$ is the eigenbase of the Kronecker product and $\vq^{(l)}$ and $\vw^{(l)}$ are vectors of eigenvalues of $\mQ_l$ and $\mW_l$ as defined before in \cref{sec:method}.
The $\diag(\cdot)$ operator turns a vector into a diagonal matrix with the corresponding vector as diagonal.
The determinant for the \ggn of one layer can be computed using properties of the Kronecker product and determinant:
\begin{align*}
  | [\mH_\param^\ggn]_l | &\approx | \mQ_l \kron \mW_l + p_\theta^{(l)} \mI_l |
  = |\mM_l (\diag(\vq^{(l)}) \kron \diag(\vw^{(l)})) \mM_l\transpose + p_\theta^{(l)} \mI_l| \\
                          &= |\mM_l (\diag(\vq^{(l)}) \kron \diag(\vw^{(l)}) + p_\theta^{(l)} \mI_l) \mM_l\transpose|
                          = | \diag(\vq^{(l)}) \kron \diag(\vw^{(l)}) + p_\theta^{(l)}| \\
                          &= \prod_{ij} \vq_i^{(l)} \vw_j^{(l)} +  p_\theta^{(l)}.
\end{align*}
To compute the final term, we only need the eigenvalues of both Kronecker factors and perform an outer product over the number of input and output neurons.
The entire determinant is simply a product over the $l$ layers due to the block-diagonal structure.
The expression is given in \cref{eq:det_kron}.

We empirically compare the proposed version to the damped version and find that it is indeed necessary to use the exact version.
The performance of the online algorithm using a damped Kronecker-factored Laplace-\ggn yields significantly worse results with negative likelihoods up to two times worse.
The accuracy suffers as well with a decrease of up to $4\%$ points.
\cref{tab:app:kron_ablation} contains the results for the same experimental setup as described in \cref{sec:experiments} and \cref{tab:imgclassification}.

\begin{table*}[h!]
  \centering
  \scriptsize
  \begin{tabular}{l l | C C C | C C C }
  \toprule
  & & \multicolumn{3}{c|}{\textbf{\textsc{kfac}}} & \multicolumn{3}{c}{\textbf{\textsc{kfac} \textit{damped}}} \\
    \textbf{Dataset} & \textbf{Model} & \textbf{accuracy} & \textbf{logLik} & \textbf{MargLik} & \textbf{accuracy} & \textbf{logLik} & \textbf{MargLik} \\ \midrule
\textbf{MNIST} & \textbf{MLP} & 98.38 \pms{0.04} & -0.053 \pms{0.002} & -0.158 \pms{0.001} & 95.89 \pms{0.11} & -0.141 \pms{0.004} & -2.199 \pms{0.008}  \\
\textbf{MNIST} & \textbf{CNN} & 99.46 \pms{0.01} & -0.016 \pms{0.001} & -0.064 \pms{0.000} & 98.78 \pms{0.09} & -0.040 \pms{0.002} & -1.022 \pms{0.017}  \\
\textbf{FMNIST} & \textbf{MLP} & 89.83 \pms{0.14} & -0.305 \pms{0.006} & -0.468 \pms{0.002} & 84.12 \pms{0.23} & -0.444 \pms{0.008} & -2.022 \pms{0.026}  \\
\textbf{FMNIST} & \textbf{CNN} & 92.06 \pms{0.10} & -0.233 \pms{0.004} & -0.401 \pms{0.001} & 89.54 \pms{0.10} & -0.296 \pms{0.003} & -2.424 \pms{0.013}  \\
\textbf{CIFAR10} & \textbf{CNN} & 80.46 \pms{0.10} & -0.644 \pms{0.010} & -0.967 \pms{0.003} & 76.38 \pms{0.50} & -0.690 \pms{0.014} & -4.232 \pms{0.068}  \\
  \bottomrule
  \end{tabular}
  \caption{Ablation of Kronecker-factored Laplace without and with \emph{damping}:
    avoiding commonly used damping improves the performance across all experiments when using the online algorithm. Damping should be avoided in this context.
  }
  \label{tab:app:kron_ablation}
\end{table*}

\section{Computational details}%
\label{app:computation}
We discuss the computational considerations necessary to apply the algorithm practically, i.e., to different hyperparameters, with different \ggn and \ef approximations, and number of hyperparameter updates $K$.
In particular, we discuss the cost of line 6 for estimating the marginal likelihood and lines 7 to 10 for updating the hyperparameters and the marginal likelihood in \cref{alg:online_marglik} for individual Laplace-\ggn and \ef approximations and choices of differentiable hyperparameters.
As detailed in \cref{sub:computational}, the other parameter settings need to be chosen to enable convergence of the marginal likelihood:
the step size $\gamma$ can be set by monitoring the convergence of the marginal likelihood and hyperparameters,
and the update frequency $F$ and number of burn-in epochs $B$ should be set to ensure enough steps for convergence while keeping the computational overhead low.
The settings $F$ and $B$ simply impact the number of marginal likelihood estimations.

Recall the neural network model introduced in \cref{sec:background}:
we have likelihood $\likelihoodhyp$ and prior $\priorhyp$, where the model $\model$ consists of hyperparameters that are differentiable in the marginal likelihood and discrete choices.
Here, we entirely focus on the hyperparameters that are differentiable in the marginal likelihood.
For both regression and classification, we work with an isotropic Gaussian prior per layer
\footnote{Except for \cref{fig:toy_regression_gridsearch} where we compare against a grid-search which is expensive already for two dimensions and high resolution.}, i.e., $\priorhyp = \prod_l \gauss (\param_l| \vzero, \delta_l\inv \mI_l)$ where $l$ denotes the layer or group of parameters;
we consider the weights and biases of each layer as individual parameter groups $l$.
Hence, the Hessian of the log prior, $\mP_\param$, is diagonal but with blocks of isotropic diagonals per layer $l$.
For regression, we use a Gaussian likelihood and optimize the observation noise variance $\sigma^2$ using the marginal likelihood, i.e., we have the likelihood $\likelihoodhyp = \gauss (\vy| \vf(\vx;\param), \sigma^2)$.
For classification, we only optimize the softmax temperature $T$ in the illustrative example (\cref{fig:toy}) and have likelihood $\likelihoodhyp = \textrm{Categorical}(\vy|\softmax(\vf(\vx;\param)/T))$ since it permits $K$ cheap iterations without requiring re-computing expensive quantities.

We use either the \ggn or \ef approximation to the marginal likelihood in \cref{eq:lap_marglik} at the current neural network parameters $\paramstar$:
\begin{equation}
  \log q(\data \given \model) = \log p(\data \given \paramstar, \model) + \log p(\paramstar \given \model) + \tfrac{P}{2} \log 2\pi - \half \log \left\vert\mH_\paramstar^{\ggn/ \ef} \right\vert.
\end{equation}
The first term, the log likelihood, can be computed based on the neural network functions $\vf(\vx;\paramstar)$ over the entire dataset in \order{NP} but an unbiased estimator in \order{MP} with minibatch $M$ is also possible\footnote{We assume a standard architecture that allows a forward pass in \order{P}}.
If we update hyperparameters of the log likelihood, for example, in regression, we further store the functions.
The second term, the log prior, can be computed based on the current neural network parameters $\paramstar$ in \order{P} for our choice of prior.
As outlined in \cref{sec:method}, the major problem, and the only part in which the approximations differ, is to compute the \ggn and \ef log determinant.
In the following, we discuss the cost of one determinant computation with the different approximations and the follow-up cost of $K$ updates and hyperparameter steps.
For classification, we deal with $\model^\partial = \{\delta_l\}$ and for regression we additionally have $\sigma^2 \in \model^\partial$.

\subsection{Full Laplace-\ggn}
For the full Laplace-\ggn, we need to pass through the training data to compute all Jacobians $\mJ_\param$ and log-likelihood Hessians $\mL_\param$ with complexity \order{NPC} and \order{NC^2}, respectively.
We can then compute the log determinant in two ways depending on $N,P,C$.
If $P < NC$, we use the standard method:
we first compute the $P\times P$ matrix in \cref{eq:ggn} in \order{P^2NC + PNC^2}.
Computing the log determinant of the resulting matrix is then \order{P^3}.
Otherwise, i.e., if $NC < P$, we use the proposed formulation in \cref{sec:method}, \cref{eq:det_ef}:
in this case the computation is \order{N^2C^2P} for construction and \order{N^3C^3} to compute the determinant.
Using multiple iterations are in general not cheaper unless we have an isotropic prior and $\mP_\param \propto \mI_P$, which is common in Bayesian deep learning.
In this case, we can start with an initial eigendecomposition in \order{P^3} or \order{N^3C^3} but then update $K$ times in \order{P} and \order{N}, respectively.

\subsection{Full Laplace-\ef}
In comparison to the full Laplace-\ggn, the \ef-based alternative scales independently of the number of outputs $C$ and therefore typically cheaper and we can trade off between $P$ and $N$ directly.
We first need to compute (and store) the individual gradients in \order{NP}.
If $P < N$, we use the standard version by computing \cref{eq:efisher} in \order{P^2N} and then the determinant in \order{P^3}.
If $N < P$, we use the matrix determinant equivalence in \cref{eq:det_ef} to reduce computation to \order{N^2P} and determinant computation to \order{N^3}.
In line with the full Laplace-\ggn, multiple iterations are in general not cheaper unless $\mP_\param \propto \mI_P$ in which case we can use an initial eigendecomposition of the same complexity and iterate cheaply in \order{P} or \order{N}.

\subsection{Kronecker-factored Laplace-\ggn or \ef}
The Kronecker-factored Laplace approximations are not only cheaper to compute and store but are also significantly more efficient in updating per-layer hyperparameters due to the per-layer block-diagonal approximation.
During the pass through the training data, the Kronecker factors are computed and stored~\citep{botev2017practical} which is \order{NP+ N(D^2 + C^2 + \sum_l H_l^2)} where $H_l$ is the size of the $l$-th hidden layer.
That is because each Kronecker factor is quadratic in the number of neurons of the input or output of some layer.
We compute the eigendecomposition of all Kronecker factors in \order{D^3 C^3 + \sum_l H_l^3} but only keep the eigenvalues necessary to compute \cref{eq:det_kron}.
Updating the per-layer prior precision and observation noise variance are only \order{P} due to \cref{eq:det_kron} and many hyperparameter updates help amortize the up-front cost.
We compute the Kronecker factors using the backpack package by \citet{dangel2019backpack}.

\subsection{Diagonal Laplace-\ggn}
The diagonal Laplace-\ggn requires backpropagation of $C\times C$ matrices $\mLambda(\vy;\vf)$~\citep{dangel2019backpack} and is therefore not as fast as the diagonal \ef.
The determinant computation, however, is only \order{P}.
Once the diagonal \ggn is obtained, hyperparameter updates are as cheap as in the Kronecker-factored case in \order{P} including updates to the determinant.
Empirically, the diagonal Laplace-\ggn is not significantly faster than the Kronecker-factored Laplace but it makes a different approximation and therefore might under some circumstances work better~(see \cref{app:exp:image_classification}).

\subsection{Diagonal Laplace-\ef}
The diagonal Laplace-\ef is cheaper and simpler to compute than the diagonal Laplace-\ggn while maintaining the same hyperparameter update complexity of \order{P}.
To compute the \ef while passing through the training data, we only need to compute individual gradients, square them element-wise, and sum them up.
Hence, this is as cheap as a pass through the training data for neural network parameters (roughly \order{NP}).
One hyperparameter update is then \order{P}.

\section{Experimental details and additional results}%
\label{app:experiments}
\begin{figure*}[t]
  \begin{subfigure}{0.5\textwidth}
    \begin{tikzpicture}
      \node (plot){\includegraphics{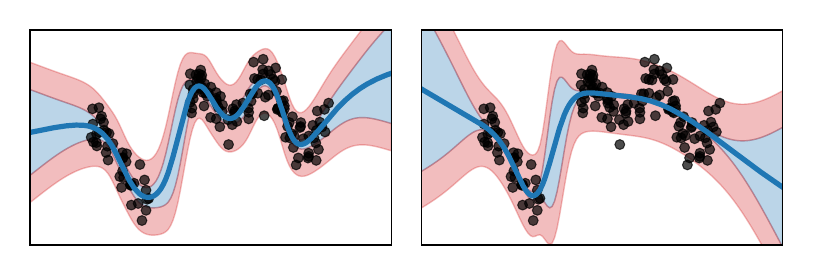}};
    \node at ($(plot.north)+(-1.9,0.0)$) {
        \begin{tikzpicture}
          \node [text width=3.5cm,align=center] {3 layers, $5221$ params MargLik = $\mathbf{-88}$};
        \end{tikzpicture}
    };
    \node at ($(plot.north)+(1.9,0.0)$) {
        \begin{tikzpicture}
          \node [text width=3cm,align=center] {1 layer, $151$ params MargLik = $-110$};
        \end{tikzpicture}
    };
    \end{tikzpicture}
    \vskip -1em
    \caption{Bayesian predictive for regression example.}
    \label{fig:app:toy_regression_bayes}
  \end{subfigure}
  \begin{subfigure}{0.5\textwidth}
    \begin{tikzpicture}
      \node (plot){\includegraphics{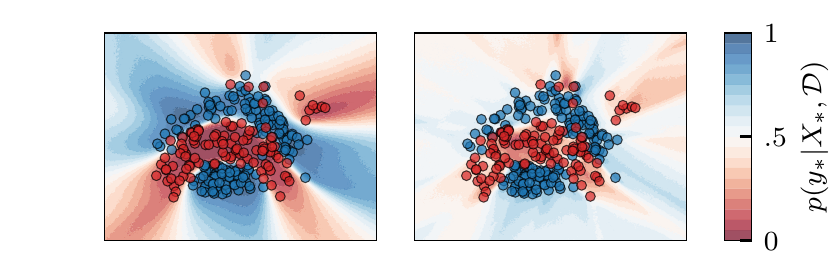}};
    \node at ($(plot.north)+(-1.8,0.0)$) {
        \begin{tikzpicture}
          \node [text width=3cm,align=center] {our method MargLik = $\mathbf{-117}$};
        \end{tikzpicture}
    };
    \node at ($(plot.north)+(1.3,0.0)$) {
        \begin{tikzpicture}
          \node [text width=3cm,align=center] {overfit\\ MargLik = $-165$};
        \end{tikzpicture}
    };
    \end{tikzpicture}
    \vskip -0.9em
    \caption{Bayesian predictive for classification example.}
    \label{fig:app:toy_classification_bayes}
  \end{subfigure}
  \caption{Optional Bayesian posterior predictives (\cref{eq:linpred}) as alternative to the MAP predictives (\cref{eq:MAPpred}) displayed in \cref{fig:toy_model_selection} and \cref{fig:toy}. (a) Posterior predictive for regression example with noise variance
    (\protect\tikz[baseline=0ex,inner sep=0pt]{\protect\draw[C4lighter, fill] (0,0.025) rectangle ++(0.5,0.15);\useasboundingbox (0,0) rectangle ++(0.5,0.15);})
  and neural network model uncertainty
  (\protect\tikz[baseline=0ex,inner sep=0pt]{\protect\draw[C1lighter, fill] (0,0.025) rectangle ++(0.5,0.15);\useasboundingbox (0,0) rectangle ++(0.5,0.15);}).
  Marginal likelihood training results in Bayesian predictives with a reasonable split into both uncertainties which is hard to obtain by other means.
  (b) The posterior predictive for classification example yields a predictive that is certain in the data region and less certain away from it when using marginal likelihood optimization.
  }
  \label{fig:app:toy_predictive}
\end{figure*}

\subsection{Illustrative examples}
\label{app:exp:illustrative}
We complement the results of \cref{sec:toy_examples} with Bayesian predictives obtained after running our online algorithm.

\paragraph{Bayesian regression predictive.}
\changed{For a Gaussian likelihood, the Bayesian predictive~(cf.~\cref{eq:linpred}) has a closed-form and we do not need samples~\citep{foong2019between, khan2019approximate, immer2020improving}.}
We additionally show the Bayesian predictive and its decomposition in epistemic and aleatoric uncertainty for the illustrative regression example in \cref{fig:toy}.
The aleatoric, irreducible, uncertainty is the noise in the data.
The epistemic uncertainty is model-dependent and quantifies how certain a model is about its prediction.
In the regression example, the aleatoric uncertainty is the observation noise $\sigma^2$ that we learn online and the epistemic uncertainty is the model uncertainty $\Var[\vf]$ which we optimize implicitly via the prior.
The variance of the posterior predictive is then $\Var[\vf] + \sigma^2$ and the standard deviation is not additive.
In \cref{fig:app:toy_regression_bayes}, we split up the total predictive variance approximately so that the observation noise $\sigma$ is depicted exactly and the model uncertainty is just the remaining standard deviation.

\paragraph{Bayesian classification predictive.}
\cref{fig:app:toy_classification_bayes} depicts the Bayesian predictives corresponding to the schematic figure \cref{fig:toy}.
\changed{We use $S=1000$ samples to estimate the posterior predictive in \cref{eq:linpred}.
The samples are on the output of the network and are therefore cheap~\citep{immer2020improving}.}
The Bayesian predictive of the model with a better marginal likelihood has increasing uncertainty away from the data as often desired~\citep{foong2019between}.
The parameters found by our online algorithm give rise to a Bayesian predictive without further tuning of parameters after training which is required when using the Laplace approximation~\citep{ritter2018scalable, kristiadi2020being}.
Alternatively, one can run a grid-search to find suitable parameters~\citep{khan2019approximate, immer2020improving}.
Our algorithm is more principled than changing the prior after training and significantly more efficient than a grid-search.

\paragraph{Regression grid search.}
\begin{wrapfigure}{r}{0.5\textwidth}
  \vskip -2em
  \includegraphics[width=0.9\linewidth]{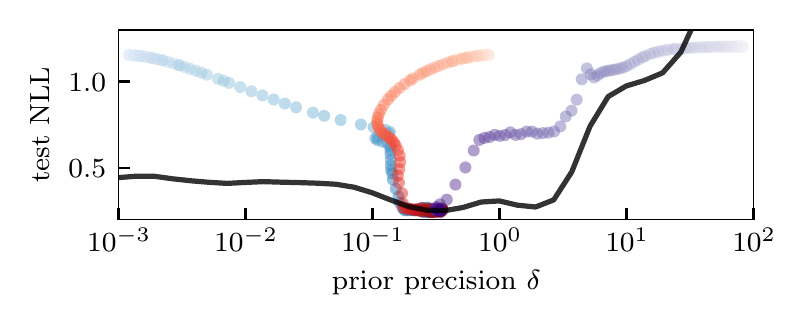}
  \vskip -1em
  \caption{Test negative log likelihood versus prior precision $\delta$ corresponding to \cref{fig:toy_regression_gridsearch}.
   The proposed method reliably converges to the optimal predictive performance.}
   \vskip -2em
\label{fig:app:regression_gridsearch}
\end{wrapfigure}
In \cref{fig:toy_regression_gridsearch}, we compared a grid-search over a single prior parameter with the online optimization algorithm and show it converges to the optimum marginal likelihood with different initializations.
Here, we additionally show in \cref{fig:app:regression_gridsearch} that the best marginal likelihood also corresponds to the best negative log likelihood (nll) on the held-out test set.
The online optimization algorithm reliably finds the optimum without access to the test data.

\subsection{UCI regression}
\label{app:exp:uci_regression}

We complement the results presented in \cref{sec:uci_regression} and in \cref{tab:uci_regression} with the performance of the optional Bayesian posterior predictive and of the diagonal determinant approximations.
We always report the performance of the posterior predictive using the posterior of the same structure as the determinant approximation during marginal likelihood optimization, e.g., if we use the full \ef during optimization, we use the Laplace-\ef posterior and make predictions with it using \cref{eq:linpred}.
In \cref{tab:app:nll_regression_bayes}, we report the performance of the additional posterior predictive for the full \ggn, \ef, and \kfac determinant computations and their corresponding posterior predictive.
The MAP predictive overall seems to perform better except on the small yacht data set.
\cref{tab:app:nll_regression_diag} further shows the performance when using the diagonal \ggn or \ef approximations.

\begin{table*}
  \centering
  \tiny
  \begin{tabular}{l | C C C | C C | C C | C C}
  \toprule
    & \multicolumn{3}{c |}{\textbf{cross-validation}} & \multicolumn{6}{c}{\textbf{MargLik optimization}} \\
\textbf{Dataset} & \textbf{MAP} & \textbf{VI} & \textbf{Laplace} & \textbf{\ggn-Bayes} & \textbf{\ggn-MAP} & \textbf{\ef-Bayes} & \textbf{\ef-MAP} & \textbf{\kfac-Bayes} & \textbf{\kfac-MAP} \\ \midrule
\textbf{boston} & 2.71 \pms{0.10} & \mathbf{2.58 \pms{0.03}} & \mathbf{2.57 \pms{0.05}} & 2.68 \pms{0.12} & 2.69 \pms{0.13} & \mathbf{2.60 \pms{0.09}} & 2.62 \pms{0.12} & 2.70 \pms{0.08} & 2.70 \pms{0.09}  \\
\textbf{concrete} & 3.17 \pms{0.04} & 3.17 \pms{0.01} & \mathbf{3.05 \pms{0.04}} & \mathbf{3.06 \pms{0.05}} & \mathbf{3.06 \pms{0.05}} & \mathbf{3.07 \pms{0.04}} & \mathbf{3.07 \pms{0.04}} & 3.13 \pms{0.06} & 3.13 \pms{0.06}  \\
\textbf{energy} & 1.02 \pms{0.05} & 1.42 \pms{0.01} & 0.82 \pms{0.03} & \mathbf{0.54 \pms{0.11}} & \mathbf{0.55 \pms{0.11}} & 0.78 \pms{0.13} & 0.73 \pms{0.15} & \mathbf{0.54 \pms{0.02}} & \mathbf{0.53 \pms{0.02}}  \\
\textbf{kin8nm} & -1.09 \pms{0.01} & -0.87 \pms{0.00} & \mathbf{-1.23 \pms{0.01}} & -1.14 \pms{0.01} & -1.14 \pms{0.01} & -1.13 \pms{0.01} & -1.14 \pms{0.01} & -1.13 \pms{0.01} & -1.13 \pms{0.01}  \\
\textbf{naval} & -5.75 \pms{0.05} & -3.82 \pms{0.02} & -6.40 \pms{0.06} & -6.61 \pms{0.03} & \mathbf{-6.93 \pms{0.03}} & -6.21 \pms{0.05} & \mathbf{-6.92 \pms{0.04}} & -6.32 \pms{0.05} & -6.88 \pms{0.05}  \\
\textbf{power} & 2.82 \pms{0.01} & 2.86 \pms{0.01} & 2.83 \pms{0.01} & \mathbf{2.78 \pms{0.02}} & \mathbf{2.78 \pms{0.02}} & \mathbf{2.78 \pms{0.01}} & \mathbf{2.78 \pms{0.01}} & \mathbf{2.78 \pms{0.02}} & \mathbf{2.78 \pms{0.02}}  \\
\textbf{wine} & 0.98 \pms{0.02} & 0.96 \pms{0.01} & 0.97 \pms{0.02} & \mathbf{0.94 \pms{0.02}} & \mathbf{0.93 \pms{0.02}} & \mathbf{0.94 \pms{0.01}} & \mathbf{0.93 \pms{0.02}} & \mathbf{0.94 \pms{0.02}} & \mathbf{0.94 \pms{0.02}}  \\
\textbf{yacht} & 2.30 \pms{0.02} & 1.67 \pms{0.01} & \mathbf{1.01 \pms{0.05}} & 3.51 \pms{0.62} & 5.89 \pms{1.25} & 1.29 \pms{0.17} & 2.43 \pms{0.61} & 1.48 \pms{0.07} & 1.48 \pms{0.07}  \\
  \bottomrule
  \end{tabular}
  \caption{Additional performance of the Bayesian posterior predictive on the UCI regression benchmark complementing \cref{tab:uci_regression}.
    The posterior predictive does not significantly improve the predictive except on the yacht data set which is very small and might require better uncertainty quantification.
  }
  \label{tab:app:nll_regression_bayes}
\end{table*}

\begin{table*}
  \centering
  \tiny
  \begin{tabular}{l | C C | C C | C C}
  \toprule
  \textbf{Dataset} & \textbf{\ggn-Bayes} & \textbf{\ggn-MAP} & \textbf{diag-\ggn-Bayes} & \textbf{diag-\ggn-MAP} & \textbf{diag-\ef-Bayes} & \textbf{diag-\ef-MAP} \\ \midrule
  \textbf{boston} & \mathbf{2.68 \pms{0.12}} & \mathbf{2.69 \pms{0.13}} & 2.84 \pms{0.05} & 2.84 \pms{0.05} & 2.90 \pms{0.03} & 2.89 \pms{0.03}  \\
  \textbf{concrete} & \mathbf{3.06 \pms{0.05}} & \mathbf{3.06 \pms{0.05}} & 3.13 \pms{0.04} & 3.13 \pms{0.04} & \mathbf{3.11 \pms{0.02}} & \mathbf{3.11 \pms{0.02}}  \\
  \textbf{energy} & \mathbf{0.54 \pms{0.11}} & \mathbf{0.55 \pms{0.11}} & 0.70 \pms{0.03} & 0.69 \pms{0.03} & 0.88 \pms{0.05} & 0.85 \pms{0.05}  \\
  \textbf{kin8nm} & \mathbf{-1.14 \pms{0.01}} & \mathbf{-1.14 \pms{0.01}} & \mathbf{-1.14 \pms{0.01}} & \mathbf{-1.13 \pms{0.01}} & \mathbf{-1.13 \pms{0.01}} & \mathbf{-1.13 \pms{0.01}}  \\
  \textbf{naval} & -6.61 \pms{0.03} & \mathbf{-6.93 \pms{0.03}} & -6.17 \pms{0.05} & \mathbf{-6.91 \pms{0.08}} & -5.83 \pms{0.06} & \mathbf{-6.95 \pms{0.04}}  \\
  \textbf{power} & \mathbf{2.78 \pms{0.02}} & \mathbf{2.78 \pms{0.02}} & \mathbf{2.79 \pms{0.02}} & \mathbf{2.79 \pms{0.02}} & \mathbf{2.79 \pms{0.01}} & \mathbf{2.79 \pms{0.01}}  \\
  \textbf{wine} & \mathbf{0.94 \pms{0.02}} & \mathbf{0.93 \pms{0.02}} & 0.96 \pms{0.01} & \mathbf{0.95 \pms{0.02}} & 0.98 \pms{0.01} & 0.96 \pms{0.02}  \\
  \textbf{yacht} & 3.51 \pms{0.62} & 5.89 \pms{1.25} & \mathbf{2.91 \pms{0.04}} & \mathbf{2.90 \pms{0.04}} & 3.16 \pms{0.02} & 3.15 \pms{0.02}  \\
  \bottomrule
  \end{tabular}
  \caption{Performance on UCI regression tasks of the diagonal \ef and \ggn online marginal likelihood optimization method compared to the full \ggn which performs best.
    The diagonal approximation slightly decreases the performance but is not significantly worse.
    In comparison to a cross-validated MAP, the cheap diagonal approximations still perform quite well (cf. first column in \cref{tab:app:nll_regression_bayes}).
  }
  \vskip -1em
  \label{tab:app:nll_regression_diag}
\end{table*}

\begin{table*}[h!t]
  \begin{subtable}[ht]{0.99\textwidth}
    \centering
    \tiny
    \begin{tabular}{l | C C C | C C | C C | C C}
    \toprule
      & \multicolumn{3}{c |}{\textbf{cross-validation}} & \multicolumn{6}{c}{\textbf{MargLik optimization}} \\
  \textbf{Dataset} & \textbf{MAP} & \textbf{VI} & \textbf{Laplace} & \textbf{\ggn-Bayes} & \textbf{\ggn-MAP} & \textbf{\ef-Bayes} & \textbf{\ef-MAP} & \textbf{\kfac-Bayes} & \textbf{\kfac-MAP} \\ \midrule
\textbf{australian} & \mathbf{0.32 \pms{0.01}} & \mathbf{0.32 \pms{0.01}} & \mathbf{0.32 \pms{0.01}} & \mathbf{0.31 \pms{0.01}} & \mathbf{0.31 \pms{0.01}} & 0.34 \pms{0.01} & \mathbf{0.31 \pms{0.01}} & \mathbf{0.32 \pms{0.01}} & \mathbf{0.31 \pms{0.01}}  \\
\textbf{cancer} & 0.13 \pms{0.03} & 0.17 \pms{0.05} & 0.10 \pms{0.01} & 0.10 \pms{0.01} & \mathbf{0.09 \pms{0.01}} & 0.25 \pms{0.01} & 0.10 \pms{0.02} & 0.10 \pms{0.01} & \mathbf{0.09 \pms{0.01}}  \\
\textbf{ionosphere} & 0.28 \pms{0.02} & \mathbf{0.23 \pms{0.02}} & 0.28 \pms{0.01} & 0.29 \pms{0.02} & 0.26 \pms{0.03} & 0.42 \pms{0.01} & 0.28 \pms{0.04} & 0.31 \pms{0.02} & 0.29 \pms{0.02}  \\
\textbf{glass} & \mathbf{0.89 \pms{0.03}} & 1.07 \pms{0.07} & \mathbf{0.88 \pms{0.01}} & 0.92 \pms{0.02} & 0.91 \pms{0.05} & 1.37 \pms{0.02} & 1.03 \pms{0.10} & 0.98 \pms{0.02} & 0.98 \pms{0.04}  \\
\textbf{vehicle} & 0.39 \pms{0.01} & \mathbf{0.37 \pms{0.01}} & 0.46 \pms{0.00} & 0.47 \pms{0.01} & 0.41 \pms{0.01} & 0.75 \pms{0.01} & \mathbf{0.38 \pms{0.01}} & 0.49 \pms{0.00} & 0.45 \pms{0.01}  \\
\textbf{waveform} & 0.36 \pms{0.00} & \mathbf{0.35 \pms{0.00}} & 0.37 \pms{0.00} & \mathbf{0.35 \pms{0.01}} & \mathbf{0.34 \pms{0.01}} & 0.43 \pms{0.01} & 0.36 \pms{0.02} & \mathbf{0.35 \pms{0.01}} & \mathbf{0.34 \pms{0.01}}  \\
\textbf{digits} & \mathbf{0.08 \pms{0.00}} & 0.14 \pms{0.01} & 0.26 \pms{0.00} & 0.31 \pms{0.01} & 0.13 \pms{0.01} & 1.81 \pms{0.01} & \mathbf{0.07 \pms{0.01}} & 0.30 \pms{0.01} & 0.16 \pms{0.01}  \\
\textbf{satellite} & \mathbf{0.23 \pms{0.00}} & 0.28 \pms{0.00} & 0.25 \pms{0.00} & 0.25 \pms{0.00} & \mathbf{0.22 \pms{0.01}} & 0.68 \pms{0.01} & 0.24 \pms{0.01} & 0.25 \pms{0.00} & 0.24 \pms{0.00}  \\
    \bottomrule
    \end{tabular}
    \caption{Negative test log likelihood (lower better)}
    \label{tab:nll1t}
\end{subtable}
\vspace{0.75em}

\begin{subtable}[ht]{0.99\textwidth}
    \centering
    \tiny
    \begin{tabular}{l | C C C | C C | C C | C C}
    \toprule
      & \multicolumn{3}{c |}{\textbf{cross-validation}} & \multicolumn{6}{c}{\textbf{MargLik optimization}} \\
  \textbf{Dataset} & \textbf{MAP} & \textbf{VI} & \textbf{Laplace} & \textbf{\ggn-Bayes} & \textbf{\ggn-MAP} & \textbf{\ef-Bayes} & \textbf{\ef-MAP} & \textbf{\kfac-Bayes} & \textbf{\kfac-MAP} \\ \midrule
\textbf{cancer} & \mathbf{96.8 \pms{0.4}} & 96.4 \pms{0.4} & \mathbf{97.0 \pms{0.3}} & \mathbf{96.8 \pms{0.4}} & \mathbf{96.9 \pms{0.4}} & \mathbf{96.7 \pms{0.5}} & \mathbf{96.7 \pms{0.5}} & \mathbf{96.7 \pms{0.5}} & \mathbf{96.7 \pms{0.5}}  \\
\textbf{ionosphere} & 90.6 \pms{0.7} & \mathbf{92.6 \pms{0.7}} & 90.6 \pms{0.6} & 90.0 \pms{1.3} & 90.0 \pms{1.3} & 90.9 \pms{1.2} & 91.5 \pms{1.0} & 89.1 \pms{1.3} & 88.9 \pms{1.4}  \\
\textbf{glass} & \mathbf{67.2 \pms{1.4}} & 63.4 \pms{1.4} & 65.3 \pms{1.4} & 64.4 \pms{2.7} & 64.1 \pms{2.8} & \mathbf{65.9 \pms{2.7}} & \mathbf{67.8 \pms{2.2}} & 63.7 \pms{2.0} & 64.7 \pms{1.9}  \\
\textbf{vehicle} & \mathbf{83.2 \pms{0.3}} & 81.4 \pms{0.4} & 80.9 \pms{0.3} & 81.7 \pms{0.9} & 82.0 \pms{0.9} & 82.1 \pms{0.6} & \mathbf{83.2 \pms{0.5}} & 80.3 \pms{0.8} & 80.3 \pms{0.8}  \\
\textbf{waveform} & \mathbf{85.7 \pms{0.3}} & \mathbf{85.6 \pms{0.3}} & \mathbf{85.7 \pms{0.3}} & \mathbf{85.9 \pms{0.6}} & \mathbf{85.9 \pms{0.5}} & 84.7 \pms{0.8} & 84.3 \pms{0.8} & \mathbf{86.0 \pms{0.5}} & \mathbf{85.9 \pms{0.6}}  \\
\textbf{digits} & \mathbf{98.1 \pms{0.1}} & 97.3 \pms{0.1} & 97.1 \pms{0.1} & 97.3 \pms{0.2} & 97.4 \pms{0.2} & 86.9 \pms{0.8} & \mathbf{98.1 \pms{0.2}} & 97.0 \pms{0.4} & 97.1 \pms{0.3}  \\
\textbf{satellite} & \mathbf{91.6 \pms{0.1}} & 89.9 \pms{0.1} & 91.3 \pms{0.1} & \mathbf{91.6 \pms{0.2}} & \mathbf{91.6 \pms{0.3}} & 91.1 \pms{0.4} & \mathbf{91.5 \pms{0.3}} & 91.0 \pms{0.2} & 91.0 \pms{0.2}  \\
    \bottomrule
    \end{tabular}
    \caption{Test accuracy (higher better)}
    \label{tab:acc1t}
\end{subtable}
\vspace{0.75em}

\begin{subtable}[ht]{0.99\textwidth}
    \centering
    \tiny
    \begin{tabular}{l | C C C | C C | C C | C C}
    \toprule
      & \multicolumn{3}{c |}{\textbf{cross-validation}} & \multicolumn{6}{c}{\textbf{MargLik optimization}} \\
  \textbf{Dataset} & \textbf{MAP} & \textbf{VI} & \textbf{Laplace} & \textbf{\ggn-Bayes} & \textbf{\ggn-MAP} & \textbf{\ef-Bayes} & \textbf{\ef-MAP} & \textbf{\kfac-Bayes} & \textbf{\kfac-MAP} \\ \midrule
\textbf{australian} & \mathbf{0.06 \pms{0.01}} & \mathbf{0.06 \pms{0.00}} & 0.07 \pms{0.01} & 0.07 \pms{0.01} & \mathbf{0.05 \pms{0.01}} & 0.09 \pms{0.01} & \mathbf{0.06 \pms{0.00}} & \mathbf{0.06 \pms{0.01}} & \mathbf{0.06 \pms{0.01}}  \\
\textbf{cancer} & \mathbf{0.03 \pms{0.00}} & \mathbf{0.03 \pms{0.00}} & \mathbf{0.03 \pms{0.00}} & \mathbf{0.03 \pms{0.00}} & \mathbf{0.03 \pms{0.00}} & 0.18 \pms{0.01} & \mathbf{0.03 \pms{0.00}} & \mathbf{0.03 \pms{0.00}} & \mathbf{0.03 \pms{0.00}}  \\
\textbf{ionosphere} & 0.08 \pms{0.00} & 0.07 \pms{0.00} & 0.11 \pms{0.00} & 0.10 \pms{0.01} & 0.08 \pms{0.01} & 0.22 \pms{0.01} & \mathbf{0.06 \pms{0.01}} & 0.10 \pms{0.01} & 0.08 \pms{0.01}  \\
\textbf{glass} & \mathbf{0.17 \pms{0.01}} & \mathbf{0.17 \pms{0.01}} & 0.19 \pms{0.01} & \mathbf{0.18 \pms{0.02}} & \mathbf{0.17 \pms{0.01}} & 0.36 \pms{0.02} & \mathbf{0.18 \pms{0.02}} & 0.19 \pms{0.02} & \mathbf{0.18 \pms{0.01}}  \\
\textbf{vehicle} & \mathbf{0.06 \pms{0.00}} & 0.07 \pms{0.00} & 0.10 \pms{0.00} & 0.12 \pms{0.01} & 0.07 \pms{0.01} & 0.30 \pms{0.01} & \mathbf{0.06 \pms{0.01}} & 0.12 \pms{0.01} & 0.08 \pms{0.01}  \\
\textbf{waveform} & 0.05 \pms{0.00} & \mathbf{0.04 \pms{0.00}} & 0.06 \pms{0.00} & 0.07 \pms{0.00} & 0.05 \pms{0.01} & 0.12 \pms{0.01} & 0.05 \pms{0.00} & 0.07 \pms{0.00} & 0.05 \pms{0.00}  \\
\textbf{digits} & \mathbf{0.01 \pms{0.00}} & 0.03 \pms{0.00} & 0.17 \pms{0.00} & 0.20 \pms{0.00} & 0.06 \pms{0.00} & 0.70 \pms{0.01} & \mathbf{0.01 \pms{0.00}} & 0.19 \pms{0.00} & 0.08 \pms{0.00}  \\
\textbf{satellite} & \mathbf{0.02 \pms{0.00}} & \mathbf{0.02 \pms{0.00}} & 0.04 \pms{0.00} & 0.05 \pms{0.00} & \mathbf{0.02 \pms{0.00}} & 0.36 \pms{0.00} & \mathbf{0.02 \pms{0.00}} & 0.05 \pms{0.00} & 0.03 \pms{0.00}  \\
    \bottomrule
    \end{tabular}
    \caption{Test expected calibration error (lower better)}
    \label{tab:ece1t}
\end{subtable}
\vspace{0.75em}

\begin{subtable}[ht]{0.99\textwidth}
    \centering
    \tiny
    \begin{tabular}{l | C C | C C | C C }
    \toprule
    \textbf{Dataset} & \textbf{\ggn-Bayes} & \textbf{\ggn-MAP} & \textbf{diag-\ggn-Bayes} & \textbf{diag-\ggn-MAP} & \textbf{diag-\ef-Bayes} & \textbf{diag-\ef-MAP} \\ \midrule
    \textbf{australian} & \mathbf{0.31 \pms{0.01}} & \mathbf{0.31 \pms{0.01}} & 0.35 \pms{0.01} & 0.33 \pms{0.01} & 0.34 \pms{0.01} & \mathbf{0.32 \pms{0.01}}  \\
    \textbf{cancer} & 0.10 \pms{0.01} & \mathbf{0.09 \pms{0.01}} & 0.12 \pms{0.01} & \mathbf{0.09 \pms{0.01}} & 0.16 \pms{0.01} & \mathbf{0.09 \pms{0.01}}  \\
    \textbf{ionosphere} & \mathbf{0.29 \pms{0.02}} & \mathbf{0.26 \pms{0.03}} & 0.38 \pms{0.01} & 0.35 \pms{0.01} & 0.37 \pms{0.01} & 0.31 \pms{0.02}  \\
    \textbf{glass} & \mathbf{0.92 \pms{0.02}} & \mathbf{0.91 \pms{0.05}} & 1.09 \pms{0.01} & 1.06 \pms{0.02} & 1.05 \pms{0.02} & 0.99 \pms{0.03}  \\
    \textbf{vehicle} & 0.47 \pms{0.01} & \mathbf{0.41 \pms{0.01}} & 0.75 \pms{0.01} & 0.68 \pms{0.01} & 0.71 \pms{0.01} & 0.59 \pms{0.01}  \\
    \textbf{waveform} & \mathbf{0.35 \pms{0.01}} & \mathbf{0.34 \pms{0.01}} & 0.41 \pms{0.01} & 0.38 \pms{0.01} & 0.40 \pms{0.01} & 0.37 \pms{0.01}  \\
    \textbf{digits} & 0.31 \pms{0.01} & 0.13 \pms{0.01} & 0.43 \pms{0.00} & 0.23 \pms{0.01} & 0.97 \pms{0.02} & \mathbf{0.09 \pms{0.01}}  \\
    \textbf{satellite} & 0.25 \pms{0.00} & \mathbf{0.22 \pms{0.01}} & 0.34 \pms{0.00} & 0.30 \pms{0.00} & 0.35 \pms{0.00} & 0.29 \pms{0.00}  \\
    \bottomrule
    \end{tabular}
    \caption{Negative test log likelihood (lower better) of diag-\ggn and diag-\ef compared to full \ggn.}
    \label{tab:nllcdiag}
\end{subtable}
  \caption{Performance of the proposed online marginal likelihood optimization method compared to cross-validation on UCI classification benchmark using a \ReLU network with 50 units and one hidden layer.
    The marginal likelihood optimization leads to comparable performance as cross-validation and the MAP typically performs better than Laplace or VI.
  The full \ef approximation performs best, followed by the \ggn and \kfac approximations.
  \cref{tab:nllcdiag} shows that the diagonal approximations perform worse, especially whem combined with a Bayesian predictive, but are still on a similar level as VI with cross-validation.
  Results within one standard error are in bold.}
\label{app:tab:uci_classification}
\end{table*}

\subsection{UCI classification}
\label{app:exp:uci_classification}

We present additional results on small-scale UCI classification data sets in \cref{app:tab:uci_classification}.
The setup is identical to that in the regression case but we use a different train/validation/test split of $70\%/15\%/15\%$ here. %
We use the same architecture as in the regression case, a single hidden layer with $50$ units and \ReLU activation and train for $10,000$ iterations until convergence with all methods.
For the grid-search we try $10$ different scalar prior precision values $\delta$ and select the best parameter on the validation set before re-training.
We compare the resulting cross-validated performance to our online algorithm with full \ggn, \ef, and \kfac approximation to the determinant in \cref{app:tab:uci_classification}.
All methods perform quite well in this benchmark and there are no obvious differences in the negative log likelihood (nll) in \cref{tab:nll1t}.
However, the cross-validated MAP and full \ef-MAP perform better than most other methods in terms of accuracy and expected calibration error.

\subsection{Image Classification: online algorithm compared to cross-validation}%
\label{app:exp:image_classification}

We provide additional details and the following additional results on the image classification experiments presented in \cref{sec:image_classification}:
the standard errors for \cref{tab:imgclassification}, the performance of the diagonal \ggn, and a comparison of different ResNet depths on CIFAR-10.

\paragraph{Architectures.}
We compare a fully-connected, convolutional, and residual neural network in our experiments.
The fully-connected network has four hidden layers with decreasing hidden layer sizes $1024, 512, 256, 128$.
For MNIST and FMNIST, this architecture has $P=1,494,154$ parameters.
As a standard convolutional neural network, we use a network of three convolutional layers followed by three fully-connected layers.
This network is used in standard benchmarks, for example the suite by \citet{schneider2018deepobs}.
On MNIST and FMNIST, this architecture has $P=892,010$ parameters and on CIFAR-10 it has $P=895,210$ parameters due to the additional color channels in the input.
On CIFAR-10, we additionally use a ResNet-20 architecture with $P=268,393$ parameters using the Fixup parameterization~\citep{zhang2019fixup}\footnote{Their implementation is publicly available with the corresponding standard settings on CIFAR-10: \url{https://github.com/hongyi-zhang/Fixup}}.
We find that increasing the depth of the ResNet up to a ResNet-110 did not improve the performance and marginal likelihood significantly as we show below in \cref{app:tab:resnet_depth}.
There, we additionally use ResNets with depths $32, 44, 56, 110$ which have $P=461,959$, $P=655,525$, $P=849,091$, and $P=1,720,138$ parameters, respectively.

\paragraph{Diagonal \ggn and standard errors.}
\cref{app:tab:imgclassification} contains the additional standard errors that were left out in the main text in \cref{tab:imgclassification} due to space constraints.
In \cref{app:tab:imgclassification_diag}, we compare the performance of our method using the diagonal \ef and the diagonal \ggn.
The latter has been left out in the image classification benchmark in the main text.
The table shows that the \ef works consistently better.
Note that the \ef is also significantly cheaper to compute and easier to implement than the other methods.

\begin{table*}[ht]
  \begin{adjustbox}{center}
  \tiny
  \begin{tabular}{l l | C C | C C C | C C C}
  \toprule
    & & \multicolumn{2}{c |}{\textbf{cross-validation}} & \multicolumn{6}{c}{\textbf{MargLik optimization}} \\
                     & & & & \multicolumn{3}{c|}{\textbf{\kfac}} & \multicolumn{3}{c}{\textbf{diagonal \ef}} \\
    \textbf{Dataset} & \textbf{Model} & \textbf{accuracy} & \textbf{logLik} & \textbf{accuracy} & \textbf{logLik} & \textbf{MargLik} & \textbf{accuracy} & \textbf{logLik} & \textbf{MargLik} \\ \midrule
    \textbf{MNIST} & \textbf{MLP} & 98.22 \pms{0.13} & -0.061 \pms{0.004} & 98.38 \pms{0.04} & -0.053 \pms{0.002} & -0.158 \pms{0.001} & 97.05 \pms{0.09} & -0.095 \pms{0.002} & -0.553 \pms{0.021}  \\
                   & \textbf{CNN} & 99.40 \pms{0.03} & \mathbf{-0.017 \pms{0.001}} & \mathbf{9.46 \pms{0.01}} & \mathbf{-0.016 \pms{0.001}} & \mathbf{-0.064 \pms{0.000}} & \mathbf{9.45 \pms{0.03}} & -0.019 \pms{0.001} & -0.134 \pms{0.001}  \\
                   \midrule
    \textbf{FMNIST} & \textbf{MLP} & 88.09 \pms{0.10} & -0.347 \pms{0.005} & 89.83 \pms{0.14} & -0.305 \pms{0.006} & -0.468 \pms{0.002} & 85.72 \pms{0.09} & -0.400 \pms{0.003} & -0.756 \pms{0.005}  \\
                    & \textbf{CNN} & 91.39 \pms{0.11} & -0.258 \pms{0.004} & \mathbf{92.06 \pms{0.10}} & \mathbf{-0.233 \pms{0.004}} & \mathbf{-0.401 \pms{0.001}} & 91.69 \pms{0.15} & \mathbf{-0.233 \pms{0.003}} & -0.570 \pms{0.003}  \\
                    \midrule
    \textbf{CIFAR10} & \textbf{CNN} & 77.41 \pms{0.06} & -0.680 \pms{0.004} & 80.46 \pms{0.10} & -0.644 \pms{0.010} & -0.967 \pms{0.003} & 80.17 \pms{0.29} & -0.600 \pms{0.010} & -1.359 \pms{0.010}  \\
                     & \textbf{ResNet} & 83.73 \pms{0.28} & -1.060 \pms{0.022} & \mathbf{86.11 \pms{0.39}} & -0.595 \pms{0.017} & \mathbf{-0.717 \pms{0.003}} & \mathbf{85.82 \pms{0.13}} & \mathbf{-0.464 \pms{0.007}} & -0.876 \pms{0.012}  \\
  \bottomrule
  \end{tabular}
  \end{adjustbox}
\caption{Standard errors for \cref{tab:imgclassification} in the main text.
  Best performances within one standard error per dataset in bold.
}
\label{app:tab:imgclassification}
\end{table*}

\begin{table*}[ht]
  \centering
  \tiny
  \begin{tabular}{l l | C C C | C C C}
  \toprule
                     & & \multicolumn{3}{c|}{\textbf{diagonal \ef}} & \multicolumn{3}{c}{\textbf{diagonal \ggn}} \\
    \textbf{Dataset} & \textbf{Model} & \textbf{accuracy} & \textbf{logLik} & \textbf{MargLik} & \textbf{accuracy} & \textbf{logLik} & \textbf{MargLik} \\ \midrule
    \textbf{MNIST} & \textbf{MLP} & 97.05 \pms{0.09} & -0.095 \pms{0.002} & -0.553 \pms{0.021} & 96.86 \pms{0.04} & -0.102 \pms{0.002} & -0.617 \pms{0.014}  \\
    & \textbf{CNN} & 99.45 \pms{0.03} & -0.019 \pms{0.001} & -0.134 \pms{0.001} & 99.35 \pms{0.02} & -0.021 \pms{0.001} & -0.193 \pms{0.001}  \\
    \midrule
    \textbf{FMNIST} & \textbf{MLP} & 85.72 \pms{0.09} & -0.400 \pms{0.003} & -0.756 \pms{0.005} & 85.36 \pms{0.21} & -0.410 \pms{0.006} & -0.844 \pms{0.016}  \\
    & \textbf{CNN} & 91.69 \pms{0.15} & -0.233 \pms{0.003} & -0.570 \pms{0.003} & 91.69 \pms{0.09} & -0.237 \pms{0.003} & -0.641 \pms{0.002}  \\
    \midrule
    \textbf{CIFAR10} & \textbf{CNN} & 80.17 \pms{0.29} & -0.600 \pms{0.010} & -1.359 \pms{0.010} & 79.98 \pms{0.46} & -0.587 \pms{0.013} & -1.568 \pms{0.010}  \\
  \bottomrule
  \end{tabular}
  \caption{Comparison of diagonal \ggn and \ef on the image classification benchmark.
    The diagonal \ef as presented in \cref{tab:imgclassification} performs slightly better on all datasets and is cheaper to compute.
  }
  \label{app:tab:imgclassification_diag}
\end{table*}

\paragraph{\kfac results for CIFAR-10 with data augmentation.}
We presented the performance of the marginal likelihood optimization using diagonal \ef on CIFAR-10 with and without data augmentation (DA) in~\cref{fig:resnet_bar}.
In \cref{app:tab:resnetbar_table}, we provide the results in terms of average performance over five random seeds and the performance of the \kfac \ggn variant.
On this benchmark, we see that the cheaper diagonal \ef performs typically slightly better while being more efficient than the \kfac variant.

\begin{table*}[h]
  \centering
  \tiny
  \begin{tabular}{l l | C C C C}
    \toprule
    \textbf{Dataset} & \textbf{Method} & \textbf{accuracy} & \textbf{logLik} & \textbf{ECE} & \textbf{OOD-AUC}\\
    \midrule
    \textbf{CIFAR-10} & \textbf{baseline} & 83.73 \pms{0.28} & -1.06 \pms{0.02} & 0.127 \pms{0.002} & 0.814 \pms{0.007}\\
    \textbf{CIFAR-10} & \textbf{MargLik diag \ef} & 85.82 \pms{0.13} & -0.46 \pms{0.01} & 0.048 \pms{0.002} & 0.901 \pms{0.004}\\
    \textbf{CIFAR-10} & \textbf{MargLik \kfac \ggn} & 86.11 \pms{0.39} & -0.60 \pms{0.02} & 0.085 \pms{0.003} & 0.882 \pms{0.008} \\
    \midrule
    \textbf{CIFAR-10 + DA} & \textbf{baseline} & 91.38 \pms{0.15} & -0.58 \pms{0.01} & 0.069 \pms{0.001} & 0.897 \pms{0.003} \\
    \textbf{CIFAR-10 + DA} & \textbf{MargLik diag \ef} & 91.18 \pms{0.20} & -0.38 \pms{0.02} & 0.054 \pms{0.003} & 0.900 \pms{0.003} \\
    \textbf{CIFAR-10 + DA} & \textbf{MargLik \kfac \ggn} & 91.26 \pms{0.12} & -0.54 \pms{0.02} & 0.068 \pms{0.001} & 0.896 \pms{0.006}\\
    \bottomrule
  \end{tabular}
  \caption{Results displayed in \cref{fig:resnet_bar} with additional results for the marginal likelihood training based on \kfac \ggn.}
  \label{app:tab:resnetbar_table}
\end{table*}

\paragraph{Deeper ResNets.}
In the main text, we only presented results for a ResNet-20.
That is because we found that a depth over $20$ for CIFAR-10 did not improve the marginal likelihood or performance noticeably.
Here, we briefly present the corresponding results with and without data augmentation.
Interestingly, the optimized marginal likelihood is largely unaffected by the almost $10$-fold increase of parameters. %
In \cref{app:tab:resnet_depth}, we show the performances, marginal likelihoods, and parameters with standard errors over five initializations.
Without data augmentation, the smaller models are preferred.
With data augmentation, the medium-sized models fare best in terms of marginal likelihood.

\begin{table*}[ht]
  \centering
  \tiny
  \begin{tabular}{l C | C C C | C C C}
    \toprule
    & & \multicolumn{3}{c|}{\textbf{no DA}} & \multicolumn{3}{c}{\textbf{with DA}} \\
    \textbf{Model} & \mathbf{P} & \textbf{accuracy} & \textbf{logLik} & \textbf{MargLik} & \textbf{accuracy} & \textbf{logLik} & \textbf{MargLik} \\
    \midrule
    \textbf{ResNet-20} & \num[group-separator={,}]{268393}  & 85.82 \pms{0.13} & -0.46 \pms{0.01} & -0.876 \pms{0.012} & 91.18 \pms{0.20} & -0.375 \pms{0.020} & -0.633 \pms{0.002}  \\
    \textbf{ResNet-32} & \num[group-separator={,}]{461959} & 85.66 \pms{0.29} & -0.46 \pms{0.01} & -0.898 \pms{0.025} & 91.63 \pms{0.22} & -0.355 \pms{0.017} & -0.623 \pms{0.008}  \\
    \textbf{ResNet-44} & \num[group-separator={,}]{655525} & 85.63 \pms{0.24} & -0.45 \pms{0.01} & -0.904 \pms{0.014} & 91.78 \pms{0.11} & -0.336 \pms{0.007} & -0.622 \pms{0.009}  \\
    \textbf{ResNet-56} & \num[group-separator={,}]{849091} & 85.82 \pms{0.17} & -0.44 \pms{0.01} & -0.932 \pms{0.010} & 91.62 \pms{0.15} & -0.344 \pms{0.012} & -0.625 \pms{0.013}  \\
    \textbf{ResNet-110} & \num[group-separator={,}]{1720138} & 86.11 \pms{0.24} & -0.44 \pms{0.01} & -0.937 \pms{0.009} & 91.85 \pms{0.13} & -0.323 \pms{0.009} & -0.638 \pms{0.009}  \\
    \bottomrule
  \end{tabular}
  \caption{ResNets of various depths trained on CIFAR-10 and their sizes, performances, and marginal likelihoods with and without data augmentation (DA).
    Without data augmentation, the smallest model achieves the highest marginal likelihood and the performances of all models are very close to each other.
    With data augmentation, the marginal likelihoods overlap in terms of their standard errors but a ResNet-44 seems best.
    Notably, the increase of the number of parameters $P$ does not significantly impact the marginal likelihood.
  }
  \label{app:tab:resnet_depth}
\end{table*}

\subsection{Image Classification: architecture selection using the marginal likelihood}%
\label{app:exp:model_selection}
We presented results on architecture selection using MargLik in \cref{sec:image_classification}, \cref{fig:teaser_marglik_accuracy}, and \cref{fig:cifar100_arch}.
\cref{fig:cifar100_arch} is the upper right of \cref{fig:teaser_marglik_accuracy} and only shows the ResNets colored by width and depth.
Here, we describe details on the experimental setup and architectures, and list more detailed results.

\paragraph{Model training.}
All models where trained using \cref{alg:online_marglik} for $250$ epochs, frequency $F=5$, $K=100$ hyperparameter updates with learning rate $\gamma=1$, and no burnin $B=0$.
We use the ResNet learning rate decay as explained in \cref{sec:experiments} for all models and train with SGD with initial learning rate of $0.01$ and a batch-size of $128$.

\paragraph{Architectures.}
On FMNIST we compare MLPs and CNNs, and on CIFAR we compare CNNs and ResNets.
For the MLPs, we use $3$ different widths $50, 200, 800$ and depth from $1$ to $5$.
The CNNs consist of up to $5$ blocks of $3 \times 3$ convolutions, followed by a ReLU operation, and MaxPooling, except in the first layer.
Instead of BatchNorm, we use the fixup parameterization~\citep{zhang2019fixup} after every convolutional layer.
For the widths, we consider widths, i.e., number of channels, from $2$ to $32$ in the first convolution.
Each following convolution uses $2\times$ more channels such that after three convolutions the number of channels is $2^3=8$ times higher.
The last layer is a fully-connected layer to the class logits.
We use ResNets of depths from $8$ to $32$ on CIFAR-10 and from $20$ to $101$ on CIFAR-100.
On CIFAR-10, more depth does not further increase or change the marginal likelihood, see \cref{app:tab:resnet_depth}.
For the width, we vary the number of channels in the first convolutional layer.
For CIFAR-10, we consider widths from $16$ to $48$ and on CIFAR-10 from $32$ to $64$.
All the architectures on the corresponding data sets are listed with the marginal likelihood, accuracy, number of parameters, width, and depth in \cref{app:tab:fmnist_arch} (FMNIST), \cref{app:tab:cifar10_arch} (CIFAR-10), and \cref{app:tab:cifar100_arch} (CIFAR-100).

\begin{figure*}[ht]
     \vspace{-0.5em}
      \begin{subfigure}{0.33\textwidth}
        \centerline{\includegraphics[scale=0.92]{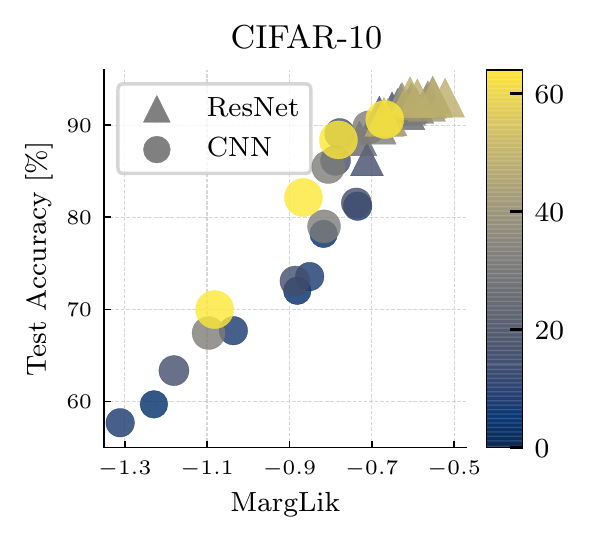}}
                    \vspace{-.5em}
      \end{subfigure}
      \begin{subfigure}{0.31\textwidth}
        \centerline{\includegraphics[scale=0.92]{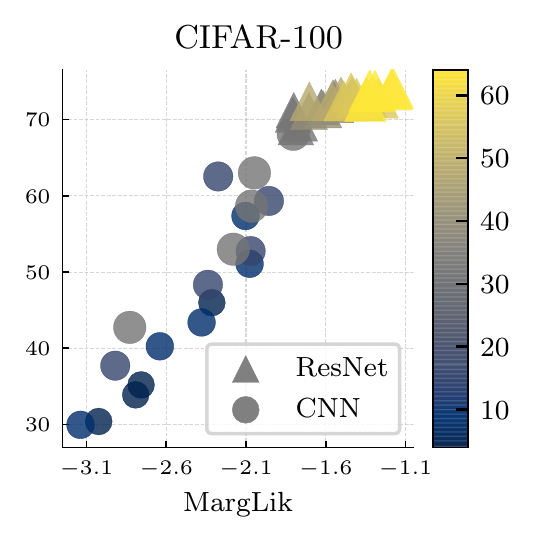}}
                    \vspace{-.5em}
      \end{subfigure}
      \begin{subfigure}{0.33\textwidth}
        \centerline{\includegraphics[scale=0.92]{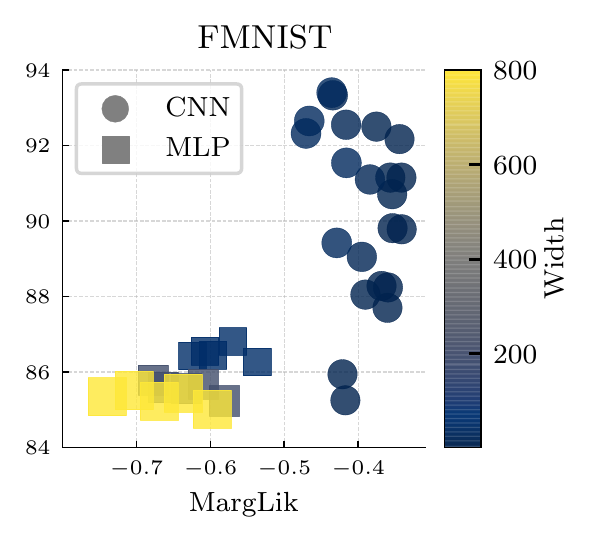}}
                    \vspace{-.5em}
      \end{subfigure}
      \vspace{-0.5em}
      \caption{Figure corresponds to \cref{fig:teaser_marglik_accuracy} but shows markers colored and scaled by \textbf{width}.}
    \label{app:fig:architectures_width}
\end{figure*}

\begin{figure*}[ht]
     \vspace{-0.5em}
      \begin{subfigure}{0.33\textwidth}
        \centerline{\includegraphics[scale=0.92]{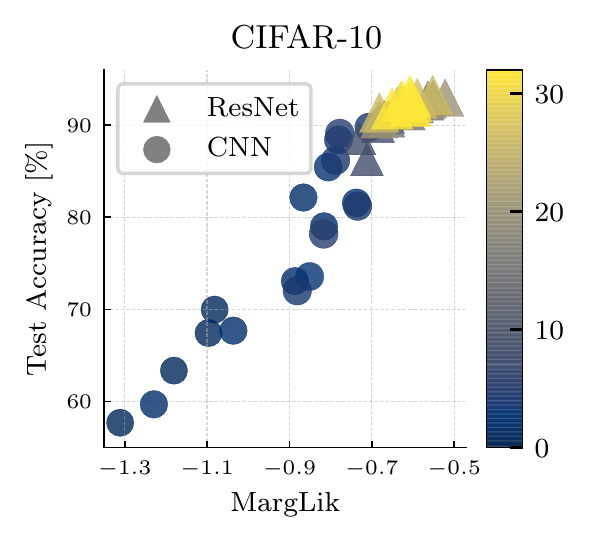}}
                    \vspace{-.5em}
      \end{subfigure}
      \begin{subfigure}{0.31\textwidth}
        \centerline{\includegraphics[scale=0.92]{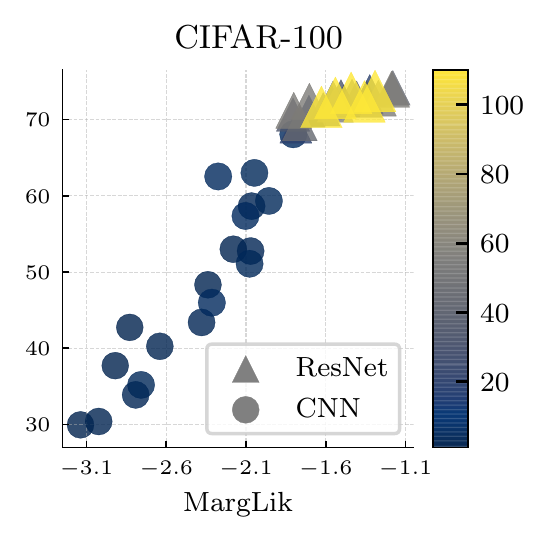}}
                    \vspace{-.5em}
      \end{subfigure}
      \begin{subfigure}{0.33\textwidth}
        \centerline{\includegraphics[scale=0.92]{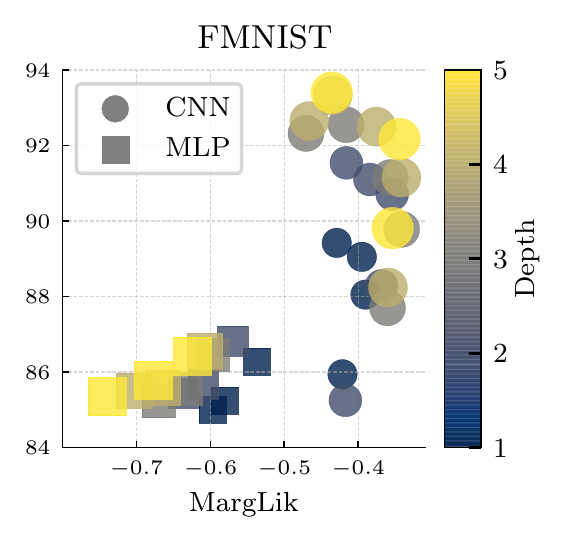}}
                    \vspace{-.5em}
      \end{subfigure}
      \vspace{-0.5em}
    \caption{Figure corresponds to \cref{fig:teaser_marglik_accuracy} but shows markers colored and scaled by \textbf{depth}.}
    \label{app:fig:architectures_depth}
\end{figure*}

\paragraph{Additional figures.}
The markers in the scatter plots of architectures are colored and scaled by the number of parameters in \cref{fig:teaser_marglik_accuracy}.
Here, we additionally show a coloring and scaling by width and depth in \cref{app:fig:architectures_width} and \cref{app:fig:architectures_depth}, respectively.
\cref{app:fig:architectures_width} shows that width tends to increase marginal likelihood and test accuracy on CIFAR.
This is in line with improved ResNet architectures that proposed increasing the width~\citep{zagoryuko2016wide}.
On FMNIST, the comparison between MLPs and CNNs is not meaningful in terms of width since width on CNNs determines the number of channels instead of hidden units.
However, we can see that less wide MLPs achieve a better marginal likelihood and test accuracy.
\cref{app:fig:architectures_depth} shows that increased depth does increase the marginal likelihood but saturates much earlier than width.
For example, ResNets, which are deeper, are generally preferred over CNNs on CIFAR but the best models are not the deepest ResNets, see also \cref{app:tab:cifar10_arch} and \cref{app:tab:cifar100_arch}.
On FMNIST, we see in \cref{app:fig:architectures_depth} how the depth tends to decrease marginal likelihood and test accuracy of MLPs.
Contrary to that, CNNs profit significantly from increasing the depth.

\subsection{Robustness to algorithm hyperparameters}%
\label{app:exp:ablation}

\cref{alg:online_marglik} has four  different hyperparameters that need to be set.
$K$ determines how many gradient steps we take on the marginal likelihood objective.
$F$ determines how frequently we compute the marginal likelihood and optimize it with respect to hyperparameters.
$B$ determines for how many epochs we do not optimize the marginal likelihood, similar to a burn-in period in MCMC algorithms.
On CIFAR-10, we conduct an ablation experiment running over a grid of these hyperparameters.
Below, we make recommendations on how to apply our algorithm based on the outcomes of this experiment.

\paragraph{Step size $\gamma$.}
The step size $\gamma$ only needs to be set to ensure the marginal likelihood objective does not diverge.
$\gamma$ does not affect the runtime but only convergence behavior.
We find that using \adam with default learning rate of $\gamma=0.001$ works well but we can increase the learning rate without divergence to $\gamma=1$ which ultimately decreases the number of steps $K$ we need to take every $F$-th epoch to observe convergence.
To set $\gamma$, a practitioner only needs to monitor the marginal likelihood or the parameters and avoid divergence or oscillation.

\paragraph{Frequency and number of steps.}
Number of steps $K$, frequency $F$, and burn-in steps $B$ can be used to reduce the runtime by avoiding to compute the Laplace approximation too often.
$B$ is only a mechanism to reduce the runtime in the early training regime where the parameters of the network $\param$ are very noisy.
If affordable, $B=0$ is optimal for performance but we find that $B=50$ works equivalently well.
In many cases, convergence of the neural network training can even be improved by adjusting hyperparameters early.
$K$ and $F$ determine how many hyperparameter steps we take and how many Laplace approximations need to be computed, respectively.
Computing the Laplace approximation is expensive while hyperparameter steps are \emph{amortized}, i.e., once we computed the Laplace approximation it is cheap to do many gradient steps $K$ (cf.~\cref{app:computation}).
It is therefore ideal for performance to have $F=1$ and compute the Laplace approximation after every epoch and have a large number of steps $K$.
In practice, however, we find that $F$ between $5$ and $10$ is sufficient and does not decrease performance.
Further, $K\geq 100$ also appears to suffice for convergence with no gain using $K=1000$.

\begin{minipage}{\textwidth}
  \centering
\begin{minipage}[b]{.43\textwidth}
  \centering
    \tiny
  \begin{tabular}{C C R R R R}
    \toprule
    \textbf{accuracy} & \textbf{MargLik} &  \multicolumn{1}{c}{\textbf{width}} &  \multicolumn{1}{c}{\textbf{depth}} &                                    \multicolumn{1}{c}{\textbf{\# params}} &
    \multicolumn{1}{c}{\textbf{model}} \\
    \midrule
    57.69 &  -1.311 &     8 &     1 &    \num[group-separator={,}]{1572} &    CNN \\
    59.69 &  -1.229 &     4 &     2 &    \num[group-separator={,}]{1238} &    CNN \\
    63.35 &  -1.180 &    16 &     1 &    \num[group-separator={,}]{3076} &    CNN \\
    67.43 &  -1.096 &    32 &     1 &    \num[group-separator={,}]{6084} &    CNN \\
    67.69 &  -1.036 &     8 &     2 &    \num[group-separator={,}]{2942} &    CNN \\
    69.98 &  -1.081 &    64 &     1 &   \num[group-separator={,}]{12100} &    CNN \\
    72.02 &  -0.881 &     4 &     4 &    \num[group-separator={,}]{7050} &    CNN \\
    73.07 &  -0.886 &    16 &     2 &    \num[group-separator={,}]{8078} &    CNN \\
    73.57 &  -0.851 &     8 &     3 &    \num[group-separator={,}]{7464} &    CNN \\
    78.20 &  -0.817 &     4 &     5 &   \num[group-separator={,}]{25588} &    CNN \\
    79.02 &  -0.816 &    32 &     2 &   \num[group-separator={,}]{25262} &    CNN \\
    81.21 &  -0.734 &     8 &     4 &   \num[group-separator={,}]{26002} &    CNN \\
    81.56 &  -0.737 &    16 &     3 &   \num[group-separator={,}]{26296} &    CNN \\
    82.15 &  -0.866 &    64 &     2 &   \num[group-separator={,}]{87278} &    CNN \\
    85.46 &  -0.806 &    32 &     3 &   \num[group-separator={,}]{98520} &    CNN \\
    86.18 &  -0.788 &    16 &     4 &  \num[group-separator={,}]{100194} &    CNN \\
    86.30 &  -0.712 &    16 &     8 &   \num[group-separator={,}]{75129} & ResNet \\
    88.41 &  -0.781 &    64 &     3 &  \num[group-separator={,}]{381208} &    CNN \\
    88.58 &  -0.730 &    24 &     8 &  \num[group-separator={,}]{167825} & ResNet \\
    89.10 &  -0.778 &    16 &     5 &  \num[group-separator={,}]{395404} &    CNN \\
    89.78 &  -0.706 &    32 &     4 &  \num[group-separator={,}]{393730} &    CNN \\
    89.90 &  -0.685 &    32 &     8 &  \num[group-separator={,}]{297385} & ResNet \\
    90.62 &  -0.662 &    16 &    14 &  \num[group-separator={,}]{172128} & ResNet \\
    90.62 &  -0.668 &    64 &     4 & \num[group-separator={,}]{1561410} &    CNN \\
    90.63 &  -0.681 &    16 &    20 &  \num[group-separator={,}]{269127} & ResNet \\
    90.89 &  -0.660 &    40 &     8 &  \num[group-separator={,}]{463809} & ResNet \\
    90.90 &  -0.670 &    48 &     8 &  \num[group-separator={,}]{667097} & ResNet \\
    91.37 &  -0.682 &    16 &    26 &  \num[group-separator={,}]{366126} & ResNet \\
    91.40 &  -0.613 &    24 &    14 &  \num[group-separator={,}]{385784} & ResNet \\
    91.83 &  -0.651 &    16 &    32 &  \num[group-separator={,}]{463125} & ResNet \\
    91.97 &  -0.613 &    32 &    20 & \num[group-separator={,}]{1071991} & ResNet \\
    92.09 &  -0.610 &    24 &    26 &  \num[group-separator={,}]{821702} & ResNet \\
    92.11 &  -0.637 &    24 &    20 &  \num[group-separator={,}]{603743} & ResNet \\
    92.12 &  -0.606 &    24 &    32 & \num[group-separator={,}]{1039661} & ResNet \\
    92.17 &  -0.593 &    32 &    14 &  \num[group-separator={,}]{684688} & ResNet \\
    92.47 &  -0.567 &    40 &    14 & \num[group-separator={,}]{1068840} & ResNet \\
    92.50 &  -0.630 &    32 &    26 & \num[group-separator={,}]{1459294} & ResNet \\
    92.62 &  -0.626 &    40 &    32 & \num[group-separator={,}]{2883933} & ResNet \\
    92.66 &  -0.604 &    32 &    32 & \num[group-separator={,}]{1846597} & ResNet \\
    92.81 &  -0.564 &    40 &    20 & \num[group-separator={,}]{1673871} & ResNet \\
    92.93 &  -0.589 &    48 &    26 & \num[group-separator={,}]{3280526} & ResNet \\
    92.98 &  -0.551 &    48 &    14 & \num[group-separator={,}]{1538240} & ResNet \\
    92.99 &  -0.522 &    48 &    20 & \num[group-separator={,}]{2409383} & ResNet \\
    93.16 &  -0.607 &    48 &    32 & \num[group-separator={,}]{4151669} & ResNet \\
    93.29 &  -0.552 &    40 &    26 & \num[group-separator={,}]{2278902} & ResNet \\
    \bottomrule
  \end{tabular}
  \captionof{table}{CIFAR-10 architectures}
  \label{app:tab:cifar10_arch}
\end{minipage}\qquad
\begin{minipage}[b]{.43\textwidth}
  \centering
    \tiny
  \begin{tabular}{C C R R R R}
    \toprule
    \textbf{accuracy} & \textbf{MargLik} &  \multicolumn{1}{c}{\textbf{width}} &  \multicolumn{1}{c}{\textbf{depth}} &                                    \multicolumn{1}{c}{\textbf{\# params}} &
    \multicolumn{1}{c}{\textbf{model}} \\
    \midrule
    23.54 &  -3.435 &     4 &     1 &     \num[group-separator={,}]{6670} &    CNN \\
    29.97 &  -3.137 &     8 &     1 &    \num[group-separator={,}]{13182} &    CNN \\
    30.41 &  -3.023 &     4 &     2 &     \num[group-separator={,}]{7808} &    CNN \\
    33.90 &  -2.791 &     4 &     3 &     \num[group-separator={,}]{8218} &    CNN \\
    35.21 &  -2.758 &     4 &     4 &    \num[group-separator={,}]{12900} &    CNN \\
    37.73 &  -2.920 &    16 &     1 &    \num[group-separator={,}]{26206} &    CNN \\
    40.27 &  -2.640 &     8 &     2 &    \num[group-separator={,}]{15992} &    CNN \\
    42.75 &  -2.829 &    32 &     1 &    \num[group-separator={,}]{52254} &    CNN \\
    43.40 &  -2.379 &     8 &     3 &    \num[group-separator={,}]{19074} &    CNN \\
    45.99 &  -2.314 &     4 &     5 &    \num[group-separator={,}]{31438} &    CNN \\
    48.34 &  -2.339 &    16 &     2 &    \num[group-separator={,}]{34088} &    CNN \\
    51.09 &  -2.078 &     8 &     4 &    \num[group-separator={,}]{37612} &    CNN \\
    52.76 &  -2.071 &    16 &     3 &    \num[group-separator={,}]{49426} &    CNN \\
    53.00 &  -2.180 &    32 &     2 &    \num[group-separator={,}]{77192} &    CNN \\
    57.36 &  -2.103 &     8 &     5 &   \num[group-separator={,}]{111510} &    CNN \\
    58.65 &  -2.065 &    32 &     3 &   \num[group-separator={,}]{144690} &    CNN \\
    59.33 &  -1.956 &    16 &     4 &   \num[group-separator={,}]{123324} &    CNN \\
    62.54 &  -2.274 &    16 &     5 &   \num[group-separator={,}]{418534} &    CNN \\
    63.01 &  -2.047 &    32 &     4 &   \num[group-separator={,}]{439900} &    CNN \\
    68.07 &  -1.804 &    32 &     5 &  \num[group-separator={,}]{1620102} &    CNN \\
    69.01 &  -1.788 &    32 &    20 &  \num[group-separator={,}]{1083601} & ResNet \\
    69.50 &  -1.763 &    32 &    44 &  \num[group-separator={,}]{2632813} & ResNet \\
    70.65 &  -1.805 &    32 &    32 &  \num[group-separator={,}]{1858207} & ResNet \\
    71.11 &  -1.706 &    40 &    20 &  \num[group-separator={,}]{1688361} & ResNet \\
    71.22 &  -1.801 &    32 &    56 &  \num[group-separator={,}]{3407419} & ResNet \\
    71.35 &  -1.616 &    40 &    32 &  \num[group-separator={,}]{2898423} & ResNet \\
    71.66 &  -1.627 &    32 &   110 &  \num[group-separator={,}]{6893146} & ResNet \\
    72.02 &  -1.540 &    40 &    56 &  \num[group-separator={,}]{5318547} & ResNet \\
    72.36 &  -1.562 &    48 &    20 &  \num[group-separator={,}]{2426753} & ResNet \\
    72.38 &  -1.358 &    48 &   110 & \num[group-separator={,}]{15493898} & ResNet \\
    72.40 &  -1.703 &    48 &    56 &  \num[group-separator={,}]{7653611} & ResNet \\
    72.45 &  -1.487 &    56 &    20 &  \num[group-separator={,}]{3298777} & ResNet \\
    72.47 &  -1.351 &    64 &    32 &  \num[group-separator={,}]{7401471} & ResNet \\
    72.78 &  -1.554 &    40 &    44 &  \num[group-separator={,}]{4108485} & ResNet \\
    72.78 &  -1.407 &    56 &    44 &  \num[group-separator={,}]{8041333} & ResNet \\
    72.82 &  -1.271 &    56 &    56 & \num[group-separator={,}]{10412611} & ResNet \\
    72.85 &  -1.540 &    40 &   110 & \num[group-separator={,}]{10763826} & ResNet \\
    72.90 &  -1.344 &    56 &    32 &  \num[group-separator={,}]{5670055} & ResNet \\
    73.02 &  -1.436 &    48 &    44 &  \num[group-separator={,}]{5911325} & ResNet \\
    73.05 &  -1.505 &    48 &    32 &  \num[group-separator={,}]{4169039} & ResNet \\
    73.51 &  -1.441 &    56 &   110 & \num[group-separator={,}]{21083362} & ResNet \\
    73.75 &  -1.291 &    64 &   110 & \num[group-separator={,}]{27532218} & ResNet \\
    73.82 &  -1.325 &    64 &    20 &  \num[group-separator={,}]{4304433} & ResNet \\
    73.98 &  -1.188 &    64 &    56 & \num[group-separator={,}]{13595547} & ResNet \\
    74.16 &  -1.180 &    64 &    44 & \num[group-separator={,}]{10498509} & ResNet \\
    \bottomrule
  \end{tabular}
  \captionof{table}{CIFAR-100 architectures}
  \label{app:tab:cifar100_arch}
\end{minipage}
\vspace{-3em}
\end{minipage}

\begin{table}
  \centering
    \tiny
  \begin{tabular}{C C R R R R}
    \toprule
    \textbf{accuracy} & \textbf{MargLik} &  \multicolumn{1}{c}{\textbf{width}} &  \multicolumn{1}{c}{\textbf{depth}} &                                    \multicolumn{1}{c}{\textbf{\# params}} &
    \multicolumn{1}{c}{\textbf{model}} \\
    \midrule
    85.01 &  -0.598 &   800 &     1 &  \num[group-separator={,}]{636034} &  MLP \\
    85.23 &  -0.670 &   800 &     3 & \num[group-separator={,}]{1917650} &  MLP \\
    85.24 &  -0.581 &   200 &     1 &  \num[group-separator={,}]{159034} &  MLP \\
    85.25 &  -0.417 &     2 &     2 &     \num[group-separator={,}]{566} &  CNN \\
    85.35 &  -0.740 &   800 &     5 & \num[group-separator={,}]{3199266} &  MLP \\
    85.44 &  -0.637 &   800 &     2 & \num[group-separator={,}]{1276842} &  MLP \\
    85.51 &  -0.703 &   800 &     4 & \num[group-separator={,}]{2558458} &  MLP \\
    85.56 &  -0.633 &   200 &     3 &  \num[group-separator={,}]{239450} &  MLP \\
    85.59 &  -0.664 &   200 &     4 &  \num[group-separator={,}]{279658} &  MLP \\
    85.69 &  -0.610 &   200 &     2 &  \num[group-separator={,}]{199242} &  MLP \\
    85.78 &  -0.677 &   200 &     5 &  \num[group-separator={,}]{319866} &  MLP \\
    85.94 &  -0.421 &     4 &     1 &     \num[group-separator={,}]{748} &  CNN \\
    86.27 &  -0.537 &    50 &     1 &   \num[group-separator={,}]{39784} &  MLP \\
    86.43 &  -0.624 &    50 &     5 &   \num[group-separator={,}]{50016} &  MLP \\
    86.45 &  -0.597 &    50 &     3 &   \num[group-separator={,}]{44900} &  MLP \\
    86.55 &  -0.608 &    50 &     4 &   \num[group-separator={,}]{47458} &  MLP \\
    86.81 &  -0.570 &    50 &     2 &   \num[group-separator={,}]{42342} &  MLP \\
    87.70 &  -0.361 &     2 &     3 &     \num[group-separator={,}]{864} &  CNN \\
    88.05 &  -0.390 &     8 &     1 &    \num[group-separator={,}]{1428} &  CNN \\
    88.24 &  -0.360 &     2 &     4 &    \num[group-separator={,}]{2074} &  CNN \\
    88.28 &  -0.369 &     4 &     2 &    \num[group-separator={,}]{1166} &  CNN \\
    89.05 &  -0.395 &    16 &     1 &    \num[group-separator={,}]{2788} &  CNN \\
    89.42 &  -0.429 &    32 &     1 &    \num[group-separator={,}]{5508} &  CNN \\
    89.78 &  -0.341 &     4 &     3 &    \num[group-separator={,}]{2296} &  CNN \\
    89.81 &  -0.354 &     2 &     5 &    \num[group-separator={,}]{6756} &  CNN \\
    90.71 &  -0.354 &     8 &     2 &    \num[group-separator={,}]{2798} &  CNN \\
    91.10 &  -0.384 &    16 &     2 &    \num[group-separator={,}]{7790} &  CNN \\
    91.15 &  -0.342 &     4 &     4 &    \num[group-separator={,}]{6978} &  CNN \\
    91.15 &  -0.357 &     8 &     3 &    \num[group-separator={,}]{7320} &  CNN \\
    91.54 &  -0.416 &    32 &     2 &   \num[group-separator={,}]{24686} &  CNN \\
    92.17 &  -0.344 &     4 &     5 &   \num[group-separator={,}]{25516} &  CNN \\
    92.32 &  -0.471 &    32 &     3 &   \num[group-separator={,}]{97944} &  CNN \\
    92.50 &  -0.376 &     8 &     4 &   \num[group-separator={,}]{25858} &  CNN \\
    92.55 &  -0.417 &    16 &     3 &   \num[group-separator={,}]{26008} &  CNN \\
    92.65 &  -0.466 &    32 &     4 &  \num[group-separator={,}]{393154} &  CNN \\
    93.33 &  -0.434 &    16 &     4 &   \num[group-separator={,}]{99906} &  CNN \\
    93.40 &  -0.436 &    32 &     5 & \num[group-separator={,}]{1573356} &  CNN \\
    \bottomrule
  \end{tabular}
  \captionof{table}{FMNIST architectures}
  \label{app:tab:fmnist_arch}
\end{table}

\end{document}